\documentclass[pdflatex,sn-basic,iicol]{sn-jnl}% Default with double column layout

\usepackage{multirow}
\usepackage{multicol}
\usepackage{makecell}
\usepackage{caption}
\usepackage{subcaption}
\usepackage{booktabs}
\usepackage{algorithm}
\usepackage{algpseudocode}

\newcommand{\wt}[1]{{\color{black}#1}}

\newcommand{\best}[1]{{\textbf{#1}}}
\providecommand{\bai}[1]{\textcolor{black}{#1}}

\newcommand{\ie}{\textit{i.e.}}
\newcommand{\eg}{\textit{e.g.}}

\jyear{2022}%

\theoremstyle{thmstyleone}%
\theoremstyle{thmstyletwo}%

\theoremstyle{thmstylethree}%

\begin{document}

\title[Article Title]{Towards Frame Rate Agnostic Multi-Object Tracking}
%Towards Frame Rate Agnostic Multi-Object Tracking
%what about open frame rates open frame rate means testing frame rates are not in training data?


\author[1]{\fnm{Weitao} \sur{Feng}}\email{weitao.feng@sydney.edu.au}

\author*[2]{\fnm{Lei} \sur{Bai}}\email{baisanshi@gmail.com}

\author[3]{\fnm{Yongqiang} \sur{Yao}}\email{soundbupt@gmail.com}

\author[3]{\fnm{Fengwei} \sur{Yu}}\email{yufengwei@sensetime.com}

\author[2,1]{\fnm{Wanli} \sur{Ouyang}}\email{wanli.ouyang@sydney.edu.au}

\affil[1]{\orgname{The University of Sydney}, \orgaddress{\postcode{2006}, \state{NSW}, \country{Australia}}}
\affil*[2]{\orgname{Shangai AI Laboratory}, \postcode{200232}, \state{Shanghai}, \country{China}}
\affil[3]{\orgname{SenseTime Group Ltd.}, \country{China}}

% \affil[2]{\orgdiv{Department}, \orgname{Organization}, \orgaddress{\street{Street}, \city{City}, \postcode{10587}, \state{State}, \country{Country}}}

% \affil[3]{\orgdiv{Department}, \orgname{Organization}, \orgaddress{\street{Street}, \city{City}, \postcode{610101}, \state{State}, \country{Country}}}


\abstract{Multi-Object Tracking (MOT) is one of the most fundamental computer vision tasks that contributes to various video analysis applications.
Despite the recent promising progress, current MOT research is still limited to a fixed sampling frame rate of the input stream. They are neither as flexible as humans nor well-matched to industrial scenarios which require the trackers to be frame rate insensitive in complicated conditions. In fact, we empirically found that the accuracy of all recent state-of-the-art trackers drops dramatically when the input frame rate changes. For a more intelligent tracking solution, we shift the attention of our research work to the problem of 
Frame Rate Agnostic MOT (FraMOT), which takes frame rate insensitivity into consideration. In this paper, we propose a Frame Rate Agnostic MOT framework with a Periodic training Scheme (FAPS) to tackle the FraMOT problem for the first time. Specifically, we propose a Frame Rate Agnostic Association Module (FAAM) that infers and encodes the frame rate information to aid identity matching across multi-frame-rate inputs, improving the capability of the learned model in handling complex motion-appearance relations in FraMOT. Moreover, the association gap between training and inference is enlarged in FraMOT because those post-processing steps not included in training make a larger difference in lower frame rate scenarios. To address it, we propose Periodic Training Scheme (PTS) to reflect all post-processing steps in training via tracking pattern matching and fusion. Along with the proposed approaches, we make the first attempt to establish an evaluation method for this new task of FraMOT. Besides providing simulations and evaluation metrics, we try to solve new challenges in two different modes, \ie, known frame rate and unknown frame rate,  aiming to handle a more complex situation. The quantitative experiments on the challenging MOT17/20 dataset (FraMOT version) have clearly demonstrated that the proposed approaches can handle different frame rates better and thus improve the robustness against complicated scenarios.}




\maketitle

\begin{figure}
    % \centering
    \includegraphics[width=2.95in]{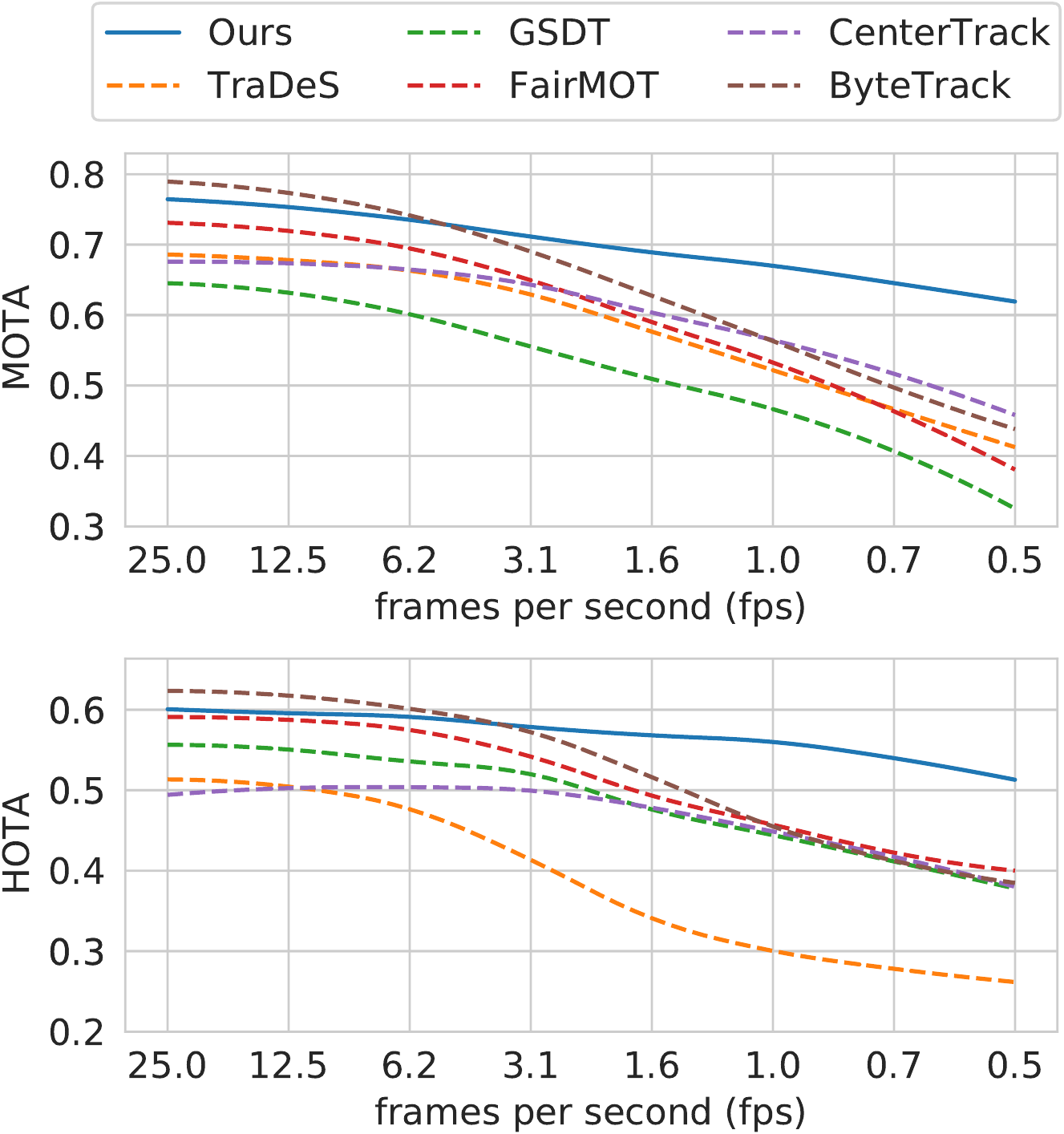}
    \caption{Performance of recent state-of-the-art trackers at multi-frame-rate settings. Both MOTA and HOTA scores drop dramatically when the frame rate is lowered. Our proposed method has a better capability of handling frame rate changes compared with previous methods. }
    %\bai{can not find frame rates in the figure}}
    \label{fig:vulnerable}
\end{figure}

\section{Introduction}\label{sec1}
%\bai{Paragraph one: significance, applications, and challenges of MOT, a few sentence}
Multi-Object Tracking (MOT) is one of the most fundamental computer vision tasks that help machines recognize the world automatically and intelligently, contributing to a great number of industrial applications of video analysis. %, \eg, large-scale trajectory retrieval, crowd counting, intelligent surveillance, and autonomous driving. 
%\bai{Paragraph two, briefly summarize existing MOT works}
Along with the development of deep learning, %MOT research at this moment is significantly different from years ago. 
%Before 2010, most MOT methods were based on motion-color-texture tracking~\cite{takala2007multi, zhang2008global}. In the 2010s, the deep learning trend and the release of MOT15-2D, MOT16, and MOT17 datasets~\cite{MOT16} brought the community to a new stage of motion-appearance-mixed tracking~\cite{chu2017online,sadeghian2017tracking} following the tracking-by-detection paradigm. After 2019, joint detection and tracking becomes popular~\cite{kieritz2018joint,zhou2020tracking,zhang2021fairmot} since it simplifies tracking procedures and reduces computational cost.
the MOT methods evolve from motion-color-texture tracking~\citep{takala2007multi, zhang2008global} to motion-appearance-mixed tracking~\citep{chu2017online,sadeghian2017tracking}, and are jointly trained with detection to simplify the procedures and reduce computational cost~\citep{kieritz2018joint,zhou2020tracking,zhang2021fairmot}.
%All these methods and datasets together contribute to the effective and efficient tracking performance.
%As the accuracy is rising, the image processing industry has started to integrate MOT algorithms in many real world applications. In return, these real world scenarios have brought back some new challenges. Among them, Frame Rate Agnostic Multi-object-tracking is one of those new challenges that is never reflected in the MOT research community.
%However, the MOT algorithms are still far from perfect, the MOT problem is nevertheless very challenging as the trackers are required to handle small targets, motion blur, frequent occlusion, and so on. On the industrial side, more challenges like limited computational capability make the task even more difficult. Thus, researchers in the MOT community are still working on MOT algorithms with better performance.

%\bai{Paragraph three: motivations for mutli-frame-rates}

%\bai{there are two aspects for Multi-frame-rate tracking: On the one hand, human has the ability to track with diverse frame rates, XXX; On the other hand, there are many practical applications working on different frame rates, training a separate tracker for each frame rate is not the best choice. First, XXX, Second, XXX, Thrid, XXX.}
Despite the aforementioned promising progress, we argue that current MOT algorithms are still not intelligent enough and cannot well satisfy industrial demands since they all work on videos with a fixed frame rate. 
%One critical problem is that current researches on MOT are all based on a fixed sampling frame rate of the input stream, which is usually an unmet condition in real world industrial scenarios. \bai{why unmet?}
%There are two aspects for multi-frame-rate tracking: 
On the one hand, humans can track objects in videos with diverse frame rates, even in a situation lacking any information on the streaming frame rate.  However, the recent State-Of-The-Art (SOTA) trackers~\citep{zhou2020tracking,wu2021track,wang2021joint,zhang2021fairmot,zhang2021bytetrack} were not robust to frame rate changes. As shown in Fig.~\ref{fig:vulnerable}, we evaluated several recent trackers (dashed lines in Fig.~\ref{fig:vulnerable})  on MOT17 and MOT20 challenges with different input frame rates and have found that their tracking capability decreases a lot when the frame rate is reduced. \wt{The ByteTrack~\citep{zhang2021bytetrack} is much better than CenterTrack~\citep{zhou2020tracking} in the normal frame rate setting but is not advanced in lower frame rate settings, showing that some tracking techniques are frame rate sensitive.} 
On the other hand, different applications require MOT models to work on different sampling rates.
Some applications (\eg, autonomous driving) require the tracker to have low latency and provide input videos of very high sampling frame rates, but more others (\eg, large-scale trajectory retrieval and tracking-based crowd counting, \cite{han2022dr}) only require the MOT algorithms to work on different lower frame rates. These applications care less about the frequent perception of object locations at every single time step but focus more on trajectory integrity along a greater temporal range. 
Although training and deploying a separate tracker for each frame rate is feasible, this trivial solution is neither convenient nor efficient because developing, selecting, and deploying the best tracker for each application and frame rate is laborious and expensive for large systems. Besides, it assumes that the frame rate is available during testing, which may not always be true.
Thus, it is necessary to propose trackers that can comprehend videos with different frame rates like humans. These trackers should be general, unified, and frame rate independent. This will be not only a reasonable objective for a smarter perception system but also an efficient solution for many real-world situations in terms of bandwidth, storage, and computation. 
%Besides, working on lower frame rates is also beneficial for bandwidth, storage, and computing resources. 
%The reasons for applying different frame rate sampling can be various, the most usual ones can be limited bandwidth and computing capabilities, strict requirements for energy efficiency. Such applications may 
%While the frame rate insensitivity is desired, we have found that 
%Currently  may be a straight-forward solution, which is, however, not the best choice. A wiser, more convenient solution would be one unified model for all frame rates. Research about how to increase robustness under multi-frame-rate settings is desired. 

%The straightforward manner to achieve frame rates agnostic MOT is  How to come up with a unified tracker for all frame rates 

%In this paper, we argue that Frame Rate Agnostic Multi-Object Tracking (FraMOT) is a necessary but overlooked branch of MOT problem, seeking for a more intelligent MOT algorithm.
%This gap causes the most MOT algorithms to be vulnerable in many industrial scenarios. 
%\bai{Paragraph Five: our method, it would be better to disentangle challenges and solution, except very hard to do so.}
To train a unified frame rate agnostic tracker, a straightforward manner is training a model of the classical design on the dataset with multiple diverse frame rates (\ie, frame rate agnostic training). However, this vanilla design does not work well due to the following two challenges.

%To address the FraMOT problem, we conducted analysis about the new challenges that it brings to us.We have identified that two key problems lead to a drop when frame rate changes. 
\textit{First}, the optimal matching rules of motion-appearance relations are different at different input frame rates. For example, motion cues are usually more reliable when the input frame rate is high, because of a smaller movement between adjacent frames. While in lower-frame-rate settings, appearance cues become more important. Two detections with similar appearances may be unlikely to be judged as the same object in higher frame rate videos given the large spatial distance but are more likely to be the same object in lower frame rate videos. Classical association models lack the mechanisms to handle these complex motion-appearance relations well.
%A straight-forward solution to solve this is to train multiple models, one for each possible frame rate,  and switch to the corresponding model during use. However, it is neither neat nor efficient. And such solution requires the exact frame rate number, which may be not feasible in some situations. Besides, it is not robust when there is a rapid camera motion, which may also causing the motion cues to be unreliable. 

\textit{Second}, involving multi-frame-rate data in traditional frame-pair association training schemes leads to a larger gap between training and inference.
\wt{Specifically, post-processing procedures that are not included in the training stage but applied in the inference stage will change the detected object locations, resulting in the input data of the association networks in the training stage being different from that in the inference stage. These changes are small in normal (higher) frame rates and thus can be ignored in the traditional training scheme. However, these changes are enlarged in the lower frame rates and are not negligible in the multi-frame-rate training. }
%Section.~\ref{sec:challenges} provides a more detailed analysis.}  
%some cases in traditional pair-wise training are not exactly the same as the inference time cases because we still have a lot post-processing in target management like trajectory status transferring, Kalman Filter, cache management and so on.} This gap becomes even bigger in frame rate agnostic training with data of diverse frame rates. The differences of data distributions make the vanilla frame rate agnostic training less effective.
%most trackers have a lot of post-processing steps (\eg, tracklet caching strategies, smoothing techniques) that are neither included in the training stage nor able to be trained in an end-to-end manner. These steps do not affect the performance obviously in higher frame rate tracking but do have great impacts when the frame rate is lower. They make the gap of input distributions between training and inference larger in FraMOT and thus make the traditional frame-pair association training scheme less effective. For instance, Kalman Filter, a commonly used motion prediction technique based on previous trajectories, predicts larger movements in lower frame rate scenarios, \wl{making the input to the association module different from the those with higher frame rates in training stage, so the knowledge learned in training stage is not completely suitable during inference.} }

\label{sec:intro}
To tackle these challenges and achieve a unified frame rate agnostic tracker, we propose the Frame Rate Agnostic MOT framework with a Periodic training Scheme (FAPS), which consists of two effective techniques. 
\textit{For the first challenge}, a unified Frame Rate Agnostic Association Module (FAAM) is proposed to handle various frame rate settings. Specifically, FAAM first employs the available frame rate information (\eg, the exact frame rate) to generate the frame rate embedding and infer the frame rate aware attention, which will further be multiplied with the prediction embedding in a channel-wise manner \wt{to predict final association scores}. For the cases where the exact frame rate is known, the frame rate embedding can be directly obtained by the frame rate cosine embedding. 
For the cases where the exact frame rate is unknown during testing, we propose to use the Inter-frame Best-matched Distance Vector (IBDV) to infer the frame rate information. 
\wt{To calculate the IBDV of two adjacent frames, we first calculate the normalized positional distance matrix, and then select the minimum distance of each row in the matrix, sorting them to form a vector. IBDV encodes the distribution of object movements and thus provides information to infer the frame rate. }
%In FAAM, the frame rate information is firstly turned into the frame rate embedding, and then generates the frame-rate-aware attention, which will be multiplied with the prediction embedding in a channel-wise manner. We further design two different frame rate embeddings for the situations of known frame rate and unknown-frame-rate, respectively.
%For the known frame rate setting, we can use the frame rate cosine embedding as the frame rate embedding, and for the unknown frame rate setting, we propose to use the Inter-frame Best-matched Distance Vector (IBDV) to infer the frame rate information. 
\textit{For the second challenge}, a Periodic Training Scheme (PTS) is designed to enhance the frame rate agnostic training. \wt{Before starting the training, we sample the tracking patterns via running previous model checkpoints on the real inference pipeline including all post-processing steps. The tracking patterns record all information we need (\ie, positions, motion predictions, and cached features) to simulate the inference stage environment during training. We assume these patterns of the tracker to have negligible variation within a short period, and thus we divide the whole training procedure into several training periods and only update patterns between periods.} During training, instances not matching those patterns will be discarded because they probably do not appear in inference time, thus reducing the difficulties of frame rate agnostic training. The remaining instances will be fused with the recorded patterns to reduce the input variance and will be turned into affinity features.
With the proposed approaches, we successfully improved the accuracy of the tracker, especially at lower frame rate settings. 
%The accuracy curve in Fig.~\ref{fig:vulnerable} shows that the proposed method has better capability of handling various input frame rates.

%\bai{Paragraph four: if we do not claim the protocol, then this paragraph would be the challenges.}
%To diminish the gap, without losing simplicity, 
To make a fair evaluation for the FraMOT task, we simulate the multi-frame-rate settings using existing benchmarking MOT datasets and develop metrics for the evaluation. Moreover, we test the methods of FraMOT task in two modes, \ie, known-frame-rate mode and unknown-frame-rate mode, reflecting different deployment demands and bringing a more challenging yet practical MOT problem to the community. Section ~\ref{sec:simulation} shows more details of the evaluation method.
%Regarding the uncertain requirements of input frame rates on deployment, we believe a good tracker is expected to handle different frame rates in a robust way, \ie, performs good enough at multiple input frame rates. 
%we first simulate the multi-frame-rate environment using the existing benchmarking MOT datasets and then use mean-HOTA~\cite{luiten2021hota} across multiple frame rates, to indicate the general performance. Similarly, we have mean-MOTA and mean-IDF1 for the classical metrics MOTA~\cite{bernardin2008evaluating} and IDF1~\cite{ristani2016performance}. We also propose a new metric named \textit{Vulnerable Ratio} (VR) to indicate the model's robustness to different frame rates, \ie, frame rate insensitivity. These metrics together provide a more accurate assessment of the trackers' performance in more complex scenarios. 
%Besides, we evaluate the trackers in two different testing modes, \ie, known-frame-rate mode and unknown-frame-rate mode. In known-frame-rate mode, the exact frame rate of the current input stream is given to the tracker. While in unknown-frame-rate mode, the only input is the image frame sequence. The two testing modes simulate different deployment demands and bring a more challenging yet practical MOT problem to the MOT community. The evaluted tracker is expected to handle multi-frame-rate input streams both with and without additional information.

In summary, our contributions are four-fold: 

\begin{itemize}
\item We, for the first time, raise the problem of Frame Rate Agnostic Multi-Object-Tracking (FraMOT), which targets on learning a unified model to track objects in videos with agnostic frame rates. Compared with the classical MOT, FraMOT is more intelligent and also practical for large vision systems.
%as an extended challenge reflecting the demand of tracking algorithms of stronger intelligence, as well as make the attempt to provide a fair evaluation method for the proposed problem. 

\item We propose a Frame Rate Agnostic MOT framework with a Periodic training Scheme (FAPS), which is the first frame rate agnostic MOT baseline attempting to effectively handle various input frame rates with a single unified model for a more robust MOT tracker in industrial scenarios. 

\item We propose a Frame Rate Agnostic Association Module (FAAM) to make use of given or inferred frame rate cues for aiding the identity matching, leading to a more intelligent tracker.

\item We propose a Periodic Training Scheme (PTS) for frame rate agnostic MOT model training, providing the simulation of inference environment and thus reducing the training-inference gap of data association.

\end{itemize}

Quantitative experiments on the challenging MOT17 and MOT20 (FraMOT versions) datasets clearly demonstrate the effectiveness of our proposed approaches at various input frame rates, which provides new insights for a more intelligent tracking solution. Code is available at \href{https://github.com/Helicopt/FraMOT}{https://github.com/Helicopt/FraMOT}.

\section{MOT in the Industry: A Context}

\subsection{Scenarios}
\noindent\textbf{Low Latency MOT.} 
In specific contexts, MOT trackers must function with minimal latency, provided that there is adequate computational capacity accessible. This classification encompasses use cases where automated assessments have the potential to result in significant outcomes. For example, real-time surveillance systems that require immediate response to potential security threats and autonomous vehicles that must react promptly to changes in their environment. In these scenarios, the ability to perform MOT with low latency can greatly enhance the effectiveness and safety of the system.

\noindent\textbf{Balanced MOT.}
In specific instances of MOT applications, certain errors may not result in catastrophic consequences. For such use cases, it is crucial that applications achieve a satisfactory level of accuracy while simultaneously maximizing their processing throughput capacity. The best approach in these scenarios is to strike a balance between processing speed and accuracy. This usually leads to the use of \textbf{low frame-rate MOT systems}. For example, in crowd monitoring applications, the goal may be to track the overall movement patterns and density of a crowd rather than identifying individual people with perfect trajectories. In traffic monitoring applications, the focus may be on detecting traffic flow and congestion rather than identifying every single vehicle. Similarly, in warehouse or logistics applications, the priority may be on tracking the overall movement of goods rather than identifying every single item with perfect accuracy. In these scenarios, a satisfactory level of accuracy coupled with high processing throughput can still provide useful insights and enable effective decision-making.

Even for applications that require a detailed path of an object, a high perception frequency may not always be necessary. For instance, in large-scale video surveillance databases, it may be sufficient to determine the location of a specific target every few seconds rather than obtaining 25 locations per second. Another example could be in sports analysis, where the movement of players and the ball needs to be tracked. However, not every frame of the video needs to be analyzed, and a lower frame rate may suffice to capture the key moments of the game.

\subsection{Robustness}
\noindent\textbf{Deployment Robustness.}
Deploying MOT algorithms is a challenging task, as the trackers involve numerous hyperparameters that must be tailored for different use cases. The frame rate is a crucial factor that impacts these hyperparameters. However, adjusting a large number of hyperparameters for a diverse range of cameras is impractical. To ensure ease of large-scale deployment, which we refer to as deployment robustness, the MOT trackers need to exhibit greater intelligence and adaptability to diverse scenarios, camera views, and frame rates.

\noindent\textbf{Hardware Robustness.}
In addition to deployment concerns, hardware-related issues are also crucial factors that can impact the robustness of MOT algorithms. For video collection hubs located far away from the data center, network issues caused by limited bandwidth or overloaded hardware capacity are unavoidable. These issues often manifest as lost packets and video lag. Furthermore, in some cases, cameras may need to be manually configured to adopt different sampling rates to accommodate temporary computational outages and recovery. All of these situations can result in dynamic changes to the sampling frame rates in the video streams. The need for hardware robustness, which involves addressing hardware-related issues, underscores the importance of frame-rate agnostic MOT algorithms. These algorithms should be capable of adapting to changes in sampling frame rates caused by hardware-related issues, while maintaining their ability to track objects robustly.

\subsection{Frame Rate Intelligence}
The variation in frame rate results in changes in the motions of objects. The challenge of handling different frame rates ultimately lies in managing different motion-appearance relationships. In some cases, the frame rate may not accurately reflect the motion distribution. For instance, a 25 fps video of a highway may have a similar motion distribution to a 1 fps pedestrian video. In such situations, using motion descriptors instead of relying solely on the frame rate may be more effective. However, in some cases, the motion descriptors may not be meaningful, such as when there are only a few objects with large covariance of motions, and the exact frame rate number may provide precise information. Therefore, we believe that an MOT solution with frame rate intelligence should be developed under two different modes: known frame rate mode and unknown frame rate mode. These modes can address the issues of different motion-appearance relationships and allow for more effective tracking.
\label{sec:test_mode}

\section{Related Works}\label{sec_related_work}

\subsection{MOT Trackers}
Most methods for MOT follow the tracking-by-detection (TBD) paradigm \citep{wojke2017simple,tang2017multiple,milan2017online,sadeghian2017tracking,chu2017online,xu2019spatial,zhou2020tracking,zhang2021fairmot}. Based on the TBD paradigm, recent MOT methods could be categorized into some different groups.

\textbf{Motion Tracking.}
Some methods~\citep{milan2017online,saleh2021probabilistic} focused on motion tracking. \cite{milan2017online} proposed to model the target motion states using RNNs and predict target motions in future frames. \cite{saleh2021probabilistic} proposed a motion model to score the tracklets, and was able to improve the performance by tracklet inpainting. 

\textbf{Complex Feature Learning.}
Other than motion features, learning appearance features is also an important part of modern MOT methods~\citep{wojke2017simple,sadeghian2017tracking,chu2017online}. \cite{wojke2017simple} proposed to borrow techniques from the Re-ID task to aid the MOT task. \cite{sadeghian2017tracking} proposed to encode motion, appearance, and interaction information for data association. \cite{chu2017online} proposed to use target-specific trackers to model the appearance of each target. 

\textbf{Joint Detection and Tracking.}
Recently joint learning of detection and tracking is also a trending branch of MOT~\citep{zhou2020tracking,zhang2021fairmot}. \cite{zhou2020tracking} proposed a point-based framework for joint detection and tracking. \cite{zhang2021fairmot} proposed to fairly balance the detection and tracking part in the joint task. Some other methods focus on the optimization of the association of trajectories. 

\textbf{Association as Graph Partition.}
While the Hungarian Algorithm is commonly used in online MOT data association, some other graph-based association methods are developed for near-online or offline MOT. For example,
\cite{tang2016multi,tang2017multiple,hornakova2020lifted} developed different graph partition methods for offline data association. \cite{yoon2019structural} further proposed to add structural constraints on the data association. \cite{hu2020dual} formulates the data association as a rank-1 tensor approximation problem. 

\textbf{Cross-Camera MOT.}
Besides single-camera MOT, \cite{wen2017multi,ma2021deep} contributed to the cross-camera MOT problem. The cross-camera MOT problem focuses on building up relationships between different cameras so that identities can be associated among multiple different cameras.

Nevertheless, all these methods have the common assumption of a fixed input frame rate, which makes the methods vulnerable in some industrial scenarios where input frame rate sampling strategies may be various. Our proposed approaches aim to establish a frame rate agnostic MOT solution that is robust to frame rate changing via unified frame rate inferring and encoding mechanisms.

\subsection{MOT Training}

There are mainly two types of MOT training, \ie, with the association in end-to-end training and without association in end-to-end training. Most recent research on joint detection and tracking separates the feature training and the association training. Typically, \cite{wojke2017simple, zhang2021fairmot} utilized deep models to perform detection and generate appearance features for tracking, and the tracker itself was not data-driven. Some hand-crafted rules were used for target association. These rules might work well in some situations, but would be difficult to be transferred to new environments. \cite{braso2020learning,li2020graph,sadeghian2017tracking,Chu_2019_ICCV} presented their methods with association in end-to-end training. However, most methods still had a post-processing stage, with hand-crafted target management. These post-processing steps were not considered in the association training. It might have minor impacts when the streaming frame rate is high and the movements of targets are small, but we found that it actually has large impacts when running a mixed frame rate training. This problem leads to a gap between association training and inference, which is even enlarged in FraMOT. The post-processing including target management and smoothing techniques like Kalman Filter also helps reduce the training difficulties. If we still follow the traditional training scheme and ignore the impacts of the post-processing, we can hardly achieve a promising tracking performance. Therefore, in this paper, we propose a new training scheme that includes the impacts from post-processing into the training stage of association, in a way of periodically updating. We thus make the training and inference environment almost the same and reduce the challenge of mixed frame rate training. \cite{maksai2019eliminating} also proposed to train a tracker in an iterative way, which aims to enrich the training data by adding wrongly tracked samples of the previous checkpoint. Our method differs from this work in two aspects: first, our method is for the online FraMOT problem and improvements are mainly from low frame rate scenarios, while ~\cite{maksai2019eliminating} works on normal frame rate multiple hypothesis tracking (MHT), which is for a totally different purpose; second, our approach is not enriching the training data but doing the opposite, aiming to reduce the gap and reduce unnecessary challenges in frame rate agnostic tracking. Our work is technically complementary to this work.

\subsection{MOT Evaluation}
CLEAR metrics~\citep{bernardin2008evaluating}, the most popular MOT evaluation metrics up to now, propose to use Multi-Object-Tracking Accuracy (MOTA) as the performance indicator for MOT. MOTA considers the False Positive (FP), False Negative (FN), and IDentity SWitch (IDSW) as three critical terms for MOT. However, the IDSW number is usually far smaller than the FN number, which makes detection quality dominate the task. Later then, the Identity metrics~\citep{ristani2016performance} and Higher Order Tracking Accuracy (HOTA)~\citep{luiten2021hota} metrics are proposed to solve such a problem. The core metric of Identity metrics, IDF1 score, is designed to evaluate MOT in the multi-camera MOT task. IDF1 basically indicates the largest ratio of consistent tracking. Due to the optimization problem in Identity metrics, IDF1 takes a long time to compute, especially when the video to be evaluated contains thousands of frames and hundreds of objects. HOTA is then proposed to solve these drawbacks of MOTA and IDF1. In the experiment of the HOTA metric, surveys were conducted and it shows that HOTA can reflect the judgments of human beings about tracking quality in a more accurate way. 

Although these metrics provide a general definition of tracking quality, it is still difficult to predict the actual performance in some complicated scenarios, \eg, a so-called better tracker may become much worse when the input frame rate changes. Unlike the change of resolution size, changing the frame rate leads to a more complicated setting which usually involves dozens of modifications of the method details. We believe that a robust tracker should have the capability of handling different frame rates. As current evaluation methods are not sufficient for frame rate agnostic MOT evaluation, we attempt to propose new evaluation methods and provide a wider analysis of tracking quality in application scenarios.

\section{Frame Rate Agnostic MOT Framework with a Periodic Training Scheme}\label{sec12}
In order to tackle the emerging challenges associated with the FraMOT problem, it is imperative to develop a novel framework that incorporates multi-frame-rate inputs and mechanisms to handle dynamic motion-appearance relationships. Additionally, the training regime of the framework should prevent an augmented gap between the training and inference stages caused by non-parametric post-processing steps under a low-frame-rate setting in the training phase.

This section presents our Frame Rate Agnostic MOT framework with a Periodic training Scheme (FAPS), which has been specifically designed to address the aforementioned objectives and overcome the challenges associated with FraMOT.

\begin{figure}
    \centering
    \includegraphics[width=3.0in]{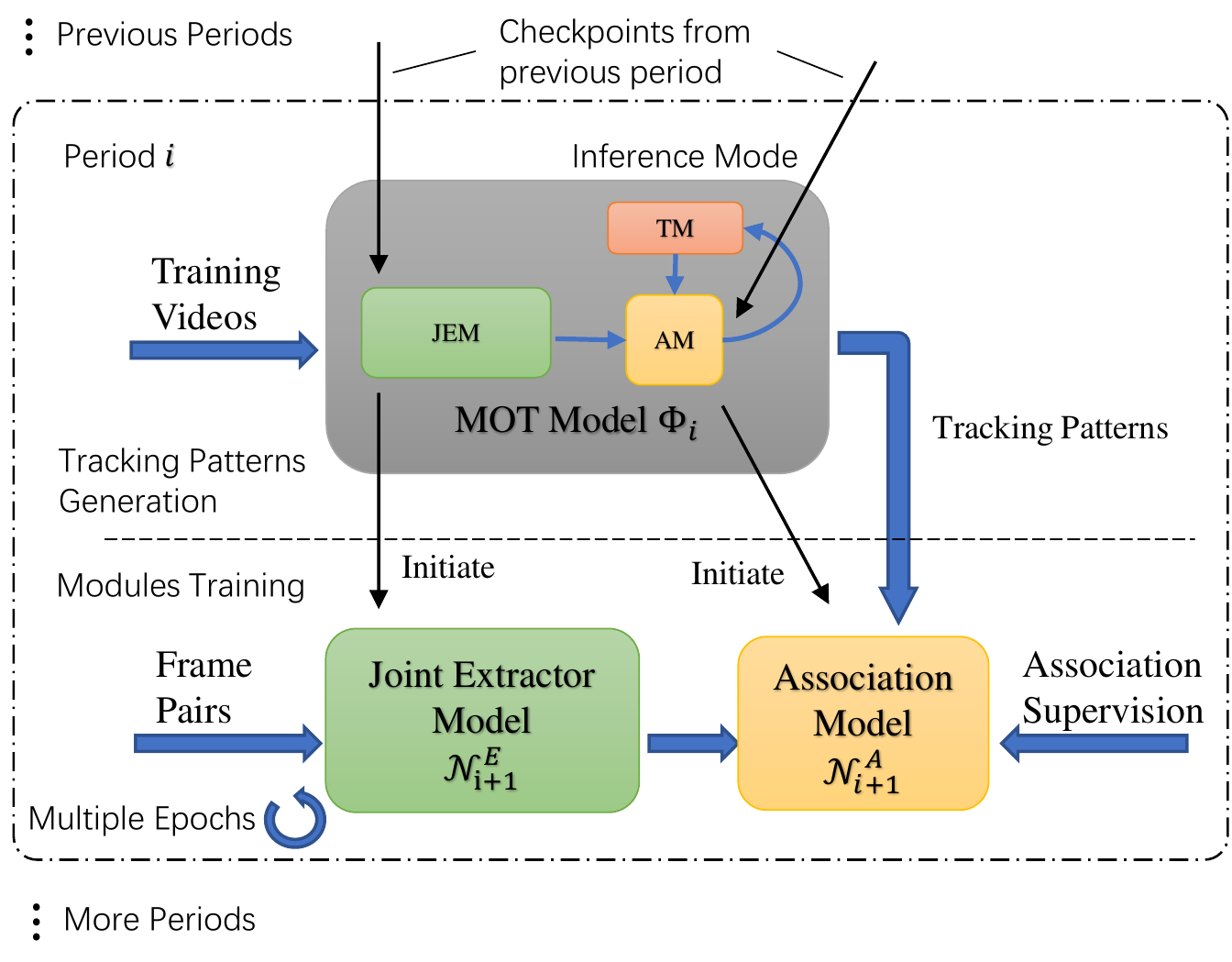}
    \caption{The training pipeline of the proposed \textbf{F}rame-rate \textbf{A}gnostic MOT framework with a \textbf{P}eriodic Training \textbf{S}cheme (FAPS) at top level. There are multiple training periods in the framework, each period contains two stages, \ie, tracking patterns generation and module training. In tracking pattern generation, we generate the tracking patterns using the previous MOT model. Then these tracking patterns will be used for module training, providing brief information on the testing environment. `JEM', `AM', and `TM' are short for Joint Extractor Module, Association Module, and Target Management, respectively.}
    \label{fig:PTS_ov}
\end{figure}

\begin{figure*}[t]
    \centering
    \includegraphics[width=6.2in]{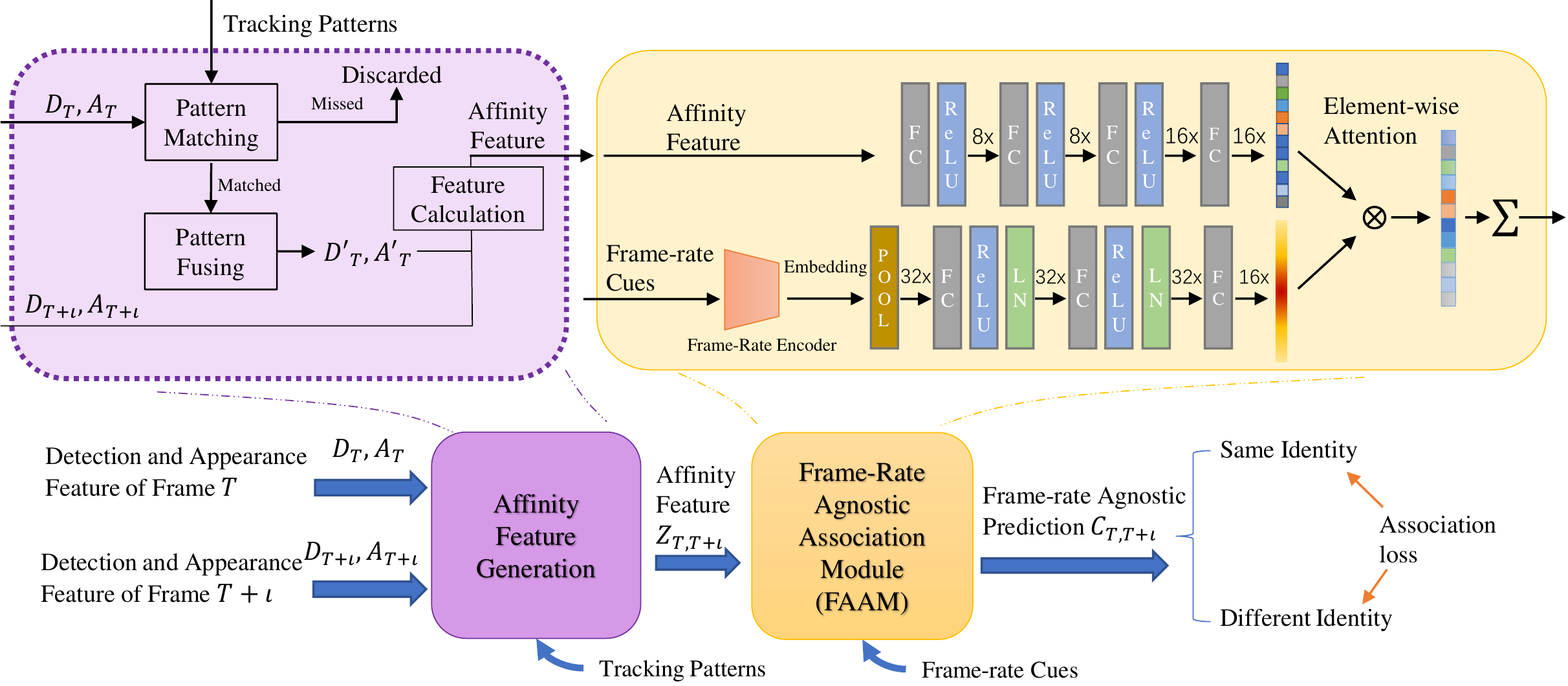}
    \caption{The training pipeline of the Association Model (AM). The detection bounding boxes $D_T, D_{T+\iota}$ and corresponding appearance features $A_T, A_{T+\iota}$ of Frame $T$ and $T+\iota$ generated by the joint extractor (\eg, YOLOX detector with an appearance embedding branch) are firstly fed to the Affinity Feature Generation Module and are matched with tracking patterns. Those matched and fused with some patterns (\ie, $D'_T$ and $A'_T$) form pairs with $D_{T+\iota} and A_{T+\iota}$ and calculate the affinity features $Z_{T,T+\iota}$. Then affinity features $Z_{T,T+\iota}$ will be fed to the Frame Rate Agnostic Association Module (FAAM) and fused with the frame rate embedding generated from the provided frame rate cues. Finally, the voted prediction $C_{T,T+\iota}$ of the module is supervised by the association loss. }
    \label{fig:pipeline_ov}
\end{figure*}

\subsection{Overview}
There are three different modules in the framework, \ie, Joint Extractor Module (JEM), Association Module (AM), and Trajectory Management Module (TM). The JEM generates the detection results and corresponding appearance feature embeddings from raw images. The AM associates the new detection results with existing trajectories. The TM determines the initiation and termination of all trajectories, \wt{making them smoother and handling their status}. 

\wt{The core module of the proposed framework is the Association Module. We design a new Frame Rate Agnostic Association Module with mechanisms to encode frame rate information, providing the capability to handle complex motion-appearance relations of the various frame rates. At the same time, the framework is trained with the proposed Periodic Training Scheme (PTS), which takes all post-processing steps into consideration, providing a simulation of the inference stage environment and thus reducing the gap of data association between training and inference. }

Fig.~\ref{fig:PTS_ov} illustrates the overview of the proposed framework. \wt{The training pipeline follows the proposed PTS consisting of several training periods, each period contains two stages, \ie, tracking patterns generation stage, and tracking module training stage. }
%\wt{We assume the inference time behaviours of the tracker remain the same within in a short period, and thus we only updates inference time tracking behaviours records between periods.} 
Specifically, the tracking patterns generation stage conducts a forward pass using the model checkpoints from the last period and generates the tracking patterns. The tracking patterns include some inference stage information that helps simulate the inference run-time (details in \ref{sec:tracking_patterns}). 
%Specially, we use a pre-trained Joint Extractor Model (JEM) and a hand-crafted Association Model (AM) to generate the tracking patterns for Period 1. 
In the module training stage, the Association Model takes the output of the JEM and the tracking patterns as input, generates the affinity features, predicts the association scores using the proposed Frame Rate Agnostic Association Module (FAAM), and is supervised by the corresponding association ground-truth signals, as shown in Fig.~\ref{fig:pipeline_ov}. Especially, during training, instead of passing the input data to the FAAM directly, we design an affinity feature generation module to adjust the affinity features utilizing the generated tracking patterns via pattern matching and fusion. Then the adjusted affinity features will be passed through the FAAM. The FAAM networks utilize the frame rate information to infer the frame rate aware attention and enhance the association prediction.
%When a training period is finished, we can utilize these checkpoints to generate the new tracking patterns for the next period. 
During Inference, the model checkpoints of the last period will be used. The association model only takes the output of the JEM as input, the tracking patterns are no longer needed and the pattern matching and fusion step is removed. The inference pipeline is the same as the tracking patterns generation pipeline.

\subsection{Periodic Training Scheme}
As mentioned before, one of the challenges that FraMOT brings to us is that the diverse video frame rates enlarge the gap between frame-pair association training and actual inference run-time. The most straightforward solution is adding these inference stage post-processing procedures to the training stage, which is hard to implement since most of these procedures can not be backward propagated. To address this issue, we propose to reflect the inference time post-processing steps during training by considering the tracking patterns in the association network (detailed in Section ~\ref{sec:affgen}), which is generated via the Periodic Training Scheme. 
%simulations of matching and fusing the affinity features with pre-generated inference stage tracking patterns from a real inference run to reflect all post-processing steps at inference time. The simulations will be detailed in Section ~\ref{sec:affgen}. 

Here, the tracking patterns are some records that can reflect the inference time patterns, which depend on the post-processing steps of the tracker.
%needed to conduct the simulations of inference stage environment during training. 
In this paper, the tracking patterns $P_t=\{(p^{loc}_i, p^{pred}_i, p^{apr}_i, p^{lvl}_i)\}$, where $  i=1,2,\ldots,N_t$, include the location of each tracked target at every frame (denoted by $p^{loc}_i$), the Kalman-Filter-predicted movement based on the previous trajectory (denoted by $p^{pred}_i$), the cached appearance feature embedding (denoted by $p^{apr}_i$), as well as the level index (denoted by $p^{lvl}_i$) of the two-stage association strategy because we use a cascade association strategy as some previous works ~\citep{yu2016poi,zhang2021fairmot}.  \label{sec:tracking_patterns} 
%The purpose of tracking patterns is to provide information for inference stage simulations during training and thus post-processing steps can be reflected in training, the design of the collection of tracking patterns will depend on the post-processing steps of the tracker. 
Other patterns such as the tracklet occlusion status can also be included if the tracker does have a carefully-designed occlusion management.

The Periodic Training Scheme aims at generating these tracking patterns in a periodic manner considering that the inference stage tracking behaviors are not going to change significantly within a short training period.
As shown in Fig.~\ref{fig:PTS_ov}, we set up multiple training periods and re-sample the tracking patterns before each period starts. In the tracking pattern generation stage of a new period, we utilize the trained checkpoints of the previous period to build an MOT tracker, conduct tracking on full training videos, and generate the tracking patterns that will be used in the affinity feature generation pipeline (see Section~\ref{sec:affgen}). Then we train all the models using the training data and tracking patterns. When a training period is done, we then have the updated checkpoints to generate tracking patterns for the next period. For the first period, we use a pre-trained joint extractor and an IoU-based hand-crafted association module to generate the patterns instead of using randomly initialized checkpoints. For example, in this paper, we use the pre-trained YOLOX detection framework~\citep{DBLP:journals/corr/abs-2107-08430} with appearance embedding branch as the JEM, a linear combination of the spatial distances and appearance similarities as the AM, and the commonly used Hungarian Algorithm and Kalman Filter with thresholding strategies for target initiation and termination as the TM (detailed in Section~\ref{sec:inference}). For the later periods, we use the proposed Frame Rate Agnostic Association Module (FAAM), and the architecture of JEM and TM remains the same.

The whole pipeline is shown in Procedure~\ref{alg:pts}. $L(\mathcal{X}, P_t;\mathcal{N}_t)$ is the overall loss which combines detection losses and id loss from the JEM, and association loss from the AM (see Section ~\ref{sec:loss}). $\lambda_E$ and $\lambda_A$ are learning rates.

\floatname{algorithm}{Procedure}
\renewcommand{\algorithmicrequire}{\textbf{Input:}}
\renewcommand{\algorithmicensure}{\textbf{Output:}}
\begin{algorithm}[t]
\caption{Periodic Training Scheme for MOT}
\label{alg:pts}
\begin{algorithmic}
\Require Period Number $N_p$, Training Data $\mathcal{X}$
\Ensure Trained Model $\Phi_{N_p}$
\State Initialize $\mathcal{N}^E_0$ with pre-trained joint extractor;
\State Initialize $\mathcal{N}^A_0$ with random parameters;
\State $\Phi_0\gets (\mathcal{N}^E_0, \gamma)$; \Comment{trivial association module $\gamma$}
\For{$t$:=$1$ to $N_p$}
    \State $P_t\gets \mathrm{Track}(\mathcal{X},\Phi_{t-1})$;
    \State $\mathcal{N}^E_t, \mathcal{N}^A_t\gets \mathcal{N}^E_{t-1}, \mathcal{N}^A_{t-1}$
    \For{$i$:=1 to $\tau$}
        \State $\mathcal{N}_t\gets \mathcal{N}^E_t, \mathcal{N}^A_t$
        \State $\mathcal{N}^E_t\gets \mathcal{N}^E_t - \lambda_E \nabla_{\mathcal{N}^E_{t-1}} L(\mathcal{X}, P_t;\mathcal{N}_t)$
        \State $\mathcal{N}^A_t\gets \mathcal{N}^A_t - \lambda_A \nabla_{\mathcal{N}^A_{t-1}} L(\mathcal{X}, P_t;\mathcal{N}_t)$
    \EndFor
    \State $\Phi_t\gets (\mathcal{N}^E_t, \mathcal{N}^A_t)$;
\EndFor
\State\Return $\Phi_{N_p}$
\end{algorithmic}
\end{algorithm}

%\textbf{Inference time tracking pattern generation.} We follow the online tracking inference time pipeline in Section~\ref{sec:inference} to generate the tracking patterns. While conducting online tracking, we collect the location of each tracked target at every frame (denoted by $p^{loc}_i$), as well as the Kalman-Filter-predicted movement based on the previous trajectory and the cached appearance feature embedding (denoted by $p^{pred}_i$ and $p^{apr}_i$, respectively). Besides we can also collect some other environment variable related to the tracking method (\eg, we can include the tracklet occlusion status if the tracker does have a carefully designed occlusion management). In our final solution, the level index (denoted by $p^{lvl}_i$) of the two-stage association strategy is also included in the affinity feature for association prediction because we use a cascade association strategy as some previous works ~\cite{yu2016poi,zhang2021fairmot}.  \label{sec:tracking_patterns}

\subsection{Frame Rate Agnostic Association}
The Association Model (AM) takes the detection results and corresponding appearance feature embeddings from the JEM and the tracking patterns (for training only) provided by the PTS as inputs and generates the affinity scores supervised by ground-truth labels. As shown in Fig.~\ref{fig:pipeline_ov}, there are two parts in the AM, \ie, Affinity Feature Generation Module and Frame Rate Agnostic Association Module.

\subsubsection{Affinity Feature Generation}
\label{sec:affgen}
\textbf{Training Stage.}
Given a frame pair of frame $T$ and $T+\iota$, where $\iota$ is \textbf{not} a constant (\ie, the sampling interval is not fixed), the JEM generates the detection results $D_T = \{B_{T,i}\mid i=1,2,\ldots,N_T\}$, $D_{T+\iota} = \{B_{T+\iota,i}\mid i=1,2,\ldots,N_{T+\iota}\}$, and the corresponding appearance features $A_{T}=\{f_{T,i}\mid i=1,2,\ldots,N_T\}$, $A_{T+\iota}=\{f_{T+\iota,i}\mid i=1,2,\ldots,N_{T+\iota}$, where $N_T$ denotes the numbers of proposals of frame $T$, $B_{T,i}$ contains the bounding box and confidence score of proposal $i$ in frame $T$, and $f_{T,i}$ represents the appearance feature vector of proposal $i$ in frame $T$, respectively.
%respectively. $D_T = \{B_{T,i}\mid i=1,2,\ldots,N_T\}$ and $D_{T+\iota} = \{B_{T+\iota,i}\mid i=1,2,\ldots,N_{T+\iota}\}$ where $B_{T,i}$ or $B_{T+\iota,i}$ is the bounding box and confidence score. Similarly we have $A_{T}=\{f_{T,i}\mid i=1,2,\ldots,N_T\}, A_{T+\iota}=\{f_{T+\iota,i}\mid i=1,2,\ldots,N_{T+\iota}$ where $f$ with dimension $D$ denotes the appearance feature vector. 
At the same time, we have the tracking patterns from frame $T$, denoted by $P_T=\{p_j\mid j=1,2,\ldots, N^{p}_T\}$, where $p_j=\{p^{loc}_j, p^{pred}_j, p^{apr}_j,p^{lvl}_j\}$ consists of a bounding box location $p^{loc}_j$, an inference stage temporal prediction $p^{pred}_j$ based on the inference environment, a saved appearance feature embedding $p^{apr}_j$ from the inference time, as well as the matching level index $p^{lvl}_j$ in the two-stage association.
We then construct a pattern matching and fusing pipeline to simulate the inference stage tracking environment for frame rate agnostic training. As shown in the left part of Fig.~\ref{fig:pipeline_ov}, only detections that are matched with a certain $p_j$ will be chosen for the training, otherwise, they will be discarded because they do not appear in the inference stage (\eg, being filtered out). In this work, we define a $B_{T,i}$ is matched with a pattern $p_j$ of the same frame if and only if $p_j$ has the largest Intersection Over Union (IoU) with $B_{T,i}$ among all available patterns, and their IoU ($\mathrm{IoU}(B_{T,i}, p^{loc}_j)$) is larger than a threshold (\eg, 0.7 as used in our work). Then the matched instances are fused with the tracking patterns, and it generates 
\begin{equation}
\begin{split}
    D'_{T} = \{&B_{T,i} + p^{pred}_{j}\mid \\
    &p_j\in P_T \land B_{T,i}\in D_T \\
    &\land p_j \mathrm{is\ matched\ with} B_{T,i}\} \\
    \cup \{&p^{loc}_{j} + p^{pred}_{j}\mid p_j\in P_T \\
    &\land p_j \mathrm{is\ \textbf{NOT} \ matched}\}.
\end{split}
\end{equation} 
The corresponding appearance features of $D'_T$ are denoted by $A'_T$. Then the corresponding embedding $f'_i\in A'_T$ of $B'_i \in D'_T$ is

\begin{equation}
\begin{split}
    f'_i = \begin{cases} f_{T,l} \quad  \mathrm{if} B'_i \mathrm{\ is\ from\ some\ } B_{T,l}, \\ p^{apr}_l \quad \mathrm{if} B'_i \mathrm{\ is\ from\ some\ } p^{loc}_l.\end{cases}
\end{split}
\end{equation} 

After matching and fusing, the $D'_T, A'_T$ and $D_{T+\iota}, A_{T+\iota}$ are then used to calculate some affinity features for the association networks. We include three distance measurements and one environment variable as the affinity features, denoted by $Z_{T,T+\iota}$, where 

\begin{equation}
\begin{split}
    Z_{T,T+\iota} = \{&\mathcal{D}(B'_i, B_{T+\iota,j}), \mathrm{IoU}(B'_i, B_{T+\iota,j}), \\
    &\frac{f'_i \cdot f_{T+\iota,j}}{\mid f'_i\mid\cdot\mid f_{T+\iota,j}\mid}, \mathrm{level}_i \\
    \mid \ &\forall B'_i \in D'_T, B_{T+\iota,j} \in D_{T+\iota} \}.
\end{split}
\end{equation}

$\mathcal{D}(\cdot)$ denotes the normalized euclidean distance and $\mathrm{level}_i$ is the matching level index $p^{lvl}_j$ in the two stage association where $B'_i$ appears in the tracking patterns.

\textbf{Inference Stage.} In the inference process, it is not mandatory to match the tracking patterns with the observations of the previous frame. The primary goal of the matching is to mimic the environment during inference time, including post-processing. However, during the actual inference stage, such a matching is unnecessary. Consequently, the generation of affinity features becomes simpler since the tracking patterns can be directly employed: 
\begin{equation}
\begin{split}
    Z_{T,T+\iota} = \{&\mathcal{D}(p_i^{loc}+p_i^{pred}, B_{T+\iota,j}), \\ 
    &\mathrm{IoU}(p_i^{loc}+p_i^{pred}, B_{T+\iota,j}), \\
    &\frac{p_i^{apr} \cdot f_{T+\iota,j}}{\mid p_i^{apr}\mid\cdot\mid f_{T+\iota,j}\mid}, \mathrm{level}_i \\
    \mid \ &\forall p_i \in P_T, B_{T+\iota,j} \in D_{T+\iota} \}.
\end{split}
\end{equation}

where $P_T$ is the tracking patterns collected from the previous frame $T$ and $D_{T+\iota}$ is the detection results of current frame $T+\iota$. To put it differently, the tracked trajectories can be viewed as cached in the tracking patterns. The association results of the tracking patterns and the new observations are exactly the association results of the tracked trajectories and the new observations.

\subsubsection{Frame Rate Agnostic Association Module} The affinity features $Z_{T,T+\iota}$ (in shape of $N_s \times D_a$, $N_s$ is the sample amount and $D_a$ is the dimension of each sample) are then fed into a 4-layer neural network $\mathcal{N}_{aff}$ and transformed to more discriminative features $f^{aff}$ (in the shape of $N_s \times 16D_a$), as shown in the right part of Fig.~\ref{fig:pipeline_ov}. 
%The channel size is then increased to 16x as the input dimension. The output of feature branch is denoted by $f^{aff}$.  
On the other branch generates the frame rate aware attention. The frame rate cues are first encoded into a frame rate embedding $\sigma$ by the frame rate encoder (see the next paragraph), then the embedding is pooled into the shape of $N_s \times 32D_a$, followed by the 3-layer frame rate sub-net $\mathcal{N}_{att}$ and finally the channel size is reduced to the same as the feature branch. The output of this branch is denoted by $f^{att}$ (in the shape of $N_s \times 16D_a$). The final output prediction of the association module is 
\begin{equation}
    C_{T,T+\iota} = \sum_{i}\frac{f^{aff}_{i}\cdot \exp(f^{att}_{i})}{\sum_j{\exp(f^{att}_{j})}}.
\end{equation}
$f^{aff}_{i}$ denotes the $i$-th element of $f^{aff}$ and $f^{att}_{j}$ denotes the $j$-th element of $f^{att}$. \label{sec:assoc}
Finally, we apply a binary cross entropy loss $L_{assoc}$ for the prediction $C_{T,T+\iota}$ with label 1 for the same identity and 0 for a different one. Specially, the label of a false detection and a true detection is 0 (they are not the same identity); the pair of two false detections will be removed from the training (unable to judge whether they are the same background region).  The overall loss $L=L_{det}+\alpha L_{id}+\beta L_{assoc}$, where $L_{det}$ and $L_{id}$ is the detection losses and id loss for the joint extractor, following the similar design of \cite{zhang2021fairmot}. \label{sec:loss}

\noindent\textbf{Frame Rate Encoder.}
%We introduce two testing modes for FraMOT, thus we introduce two different methods to calculate and generate the frame rate embedding $\sigma$ for the known frame rate mode and unknown frame rate mode, respectively. 
\bai{As discussed in Section \ref{sec:test_mode}, the frame rate could either be known or unknown during the deployment stage. Thus, two frame rate encoders are introduced correspondingly. }

\textit{For known frame rate}, we simply use the cosine embedding of the given frame rate.  Specifically, 
%we build the cosine embedding of dimension $D_\sigma$ with
the $i$-th element of the frame rate embedding $\sigma_i=cos(\frac{i\cdot s\cdot F}{D_\sigma}$), where $F$ is the frame rate number,  $s$ is a constant scaling factor, and $D_\sigma$ is the embedding dimension.

\begin{figure}
    \centering
    \includegraphics[width=2.4in]{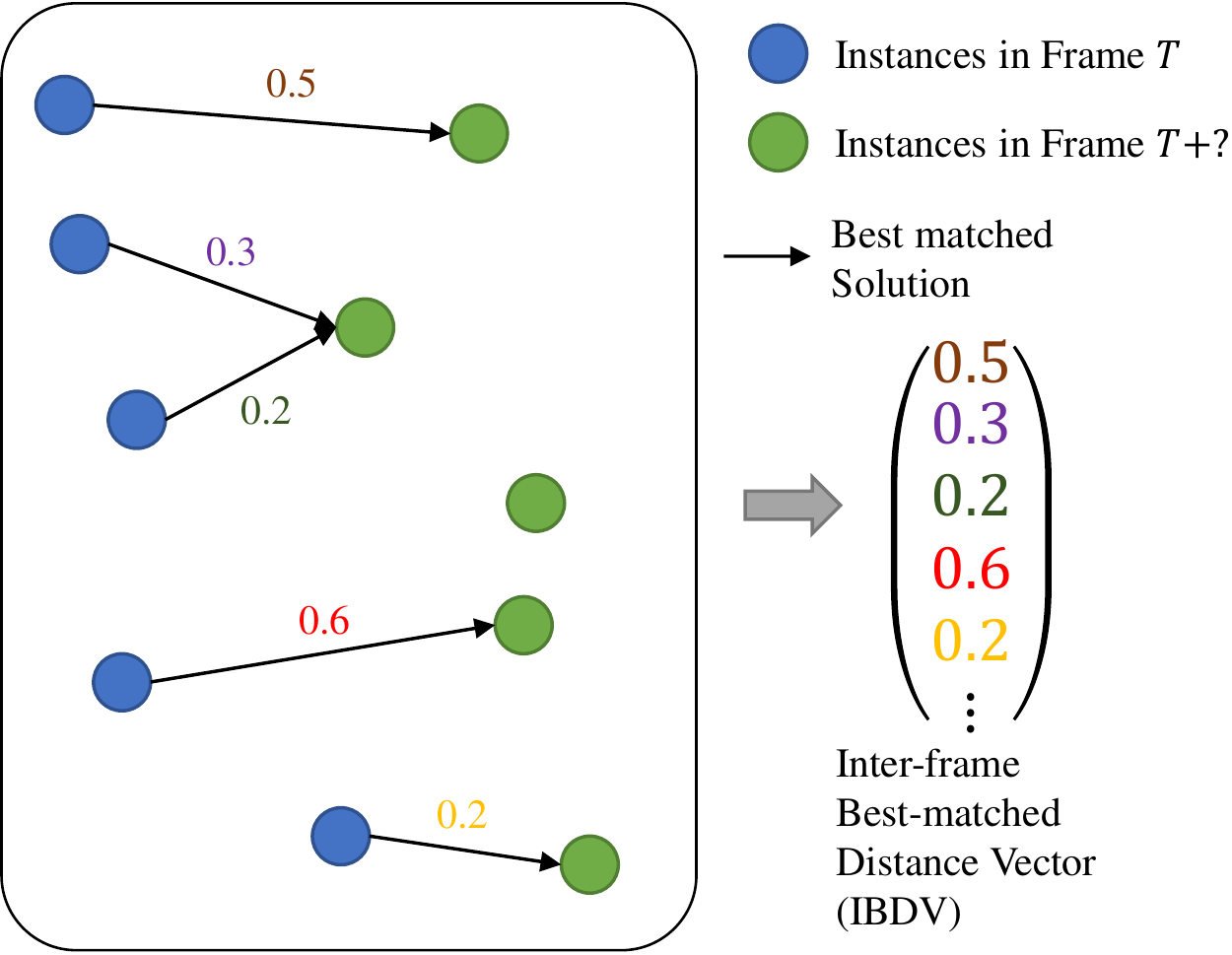}
    \caption{Illustration of Inter-frame Best-matched Distance Vector (IBDV). We calculate the normalized distances of the best matching pairs and generate the IBDV.}
    \label{fig:interframe}
\end{figure}

\label{sec:ibdv}
\textit{For unknown frame rate}, we propose to utilize the Inter-frame Best-matched Distance Vector (IBDV) as the frame rate indicator. Specifically, IBDV describes the distribution of the normalized distances between instances in the two frames if the instances are best matched. \wt{This distribution roughly encodes the frame rate information, because best-matched distances will increase when the frame rate is reduced.} As shown in Fig.~\ref{fig:interframe}, we first find the best matching object pairs through some pre-defined criteria, \eg, minimizing spatial distances or appearance feature distances, and then put all normalized distance values of these pairs into a vector. To make its dimension unchanged, we sort the values in the vector and interpolate them into a fixed shape. We investigate the effects of different criteria to find the best matching pairs in \ref{sec:bestmatching}. In our final solution, according to the results on MOT datasets, we choose the `spatial distances" as our criterion, \ie, we choose the pairs minimizing the sum of the spatial distances as our best matching pairs. Mathematically, when the frame rate is unknown, 
\begin{equation}
    \sigma = \mathrm{interp}(\mathrm{sort}(\{\mathcal{D}(B'_i, B_{T+\iota,j_i}), i=1,2,\ldots,N'\}))
\end{equation}
where $B'_i, B_{T+\iota,j_i}$ is the $i$-th matched pair, \wt{$\mathrm{interp}$ denotes the linear interpolation and $\mathrm{sort}$ denotes sorting the values of a vector in non-decreasing order}.
%Then the $\sigma$ will be used in the FAAM mentioned in Section~\ref{sec:assoc}.

\begin{algorithm}[t]
\caption{Target Management Cycle}
\label{alg:cycle}
\begin{algorithmic}
\Require Video Frames Set $I={I_1,I_2,\ldots,I_T}$
\Ensure Trajectories Set $\mathcal{T}={o_1,o_2,\ldots,o_{N_\mathcal{T}}}$
\State $\mathcal{T}\gets \emptyset$;
\For{$t$:=$1$ to $T$}
    \State $D_t, A_t\gets \mathcal{N}^E(I_t)$;
    \State $D_t, A_t\gets \mathrm{Filter}(D_t, A_t, \lambda_{low})$;
    \State $D^{high}_t, A^{high}_t \gets \mathrm{Filter}(D_t, A_t, \lambda_{high})$;
    \State $D^{low}_t, A^{low}_t \gets D_t - D^{high}_t, A_t - A^{high}_t$;
    \State $\mathcal{T}'\gets \mathrm{KalmanFilter}(\mathcal{T})$;
    \State $\mathcal{T}_{1}, D^{rem}_t, A^{rem}_t\gets $\\ $\mathrm{Match_1}(\mathcal{T}', D^{high}_t, A^{high}_t;\mathcal{N}^A)$;
    \State $\mathcal{T}_{2}\gets \mathrm{Match_2}(\mathcal{T}' - \mathcal{T}_{1}, D^{low}_t, A^{low}_t;\mathcal{N}^A)$;
    \State $T_{rem}\gets \mathrm{InitTrack}(D^{rem}_t, A^{rem}_t)$
    \State $\mathcal{T}\gets \mathcal{T}_{1} \cup \mathcal{T}_{2} \cup (\mathcal{T}'- \mathcal{T}_{1} \cup \mathcal{T}_{2}) \cup \mathcal{T}_{rem}$;
    \State $\mathcal{T}\gets \mathrm{RemoveMissing}(\mathcal{T}, \lambda_i)$;
    \State \Comment{PTS collects $p^{loc}_i, p^{pred}_i, p^{apr}_i$ from $\mathcal{T}$}
\EndFor
\State\Return $\mathcal{T}$
\end{algorithmic}
\end{algorithm}

\subsection{Online Tracking}
\label{sec:inference}
For online tracking, we adopt a simple but effective pipeline that consists of steps including detection filtering, Kalman-Filter-based movement prediction, and a two-stage association strategy. The target management cycle is stated in Procedure~\ref{alg:cycle}. We first filter out background objects with a threshold $\lambda_{low}$, then predict the movement for each tracked target using Kalman-Filter. In the next, we perform two different stages of association, \wt{which is a commonly used technique to reduce fragments~\citep{yu2016poi,zhang2021fairmot}}. For the first stage, only detection results of high confidence (confidence score greater than $\lambda_{high}$) will be put into the matching pool. For the second stage, the remaining unmatched tracked targets will be again put into the matching pool and matched with the less confident detection results (confidence score less than $\lambda_{high}$ but greater than $\lambda_{low}$). The association step will find out the best matching pairs minimizing the sum of pair distances (or maximizing the sum of pair scores). This can be easily calculated by the Hungarian Algorithm. Finally, the remaining unmatched detection results of high confidence (confidence score greater than $\lambda_{high}$) will be appended to the tracked target list. If a target in the tracked target list is unmatched for a long interval $\lambda_i$, then it will be dropped from the list.

\section{Experiment}\label{sec13}
In this section, we provide quantitative experiment results, ablation studies, and in-deep analysis of the proposed approaches to tackle the FraMOT problem.

%\subsection{Multi-Frame-Rate Simulation}
\subsection{Evaluation Datasets and Metrics}

\subsubsection{Datasets}
\label{sec:simulation}
\begin{figure}
    \centering
    \includegraphics[width=3.0in]{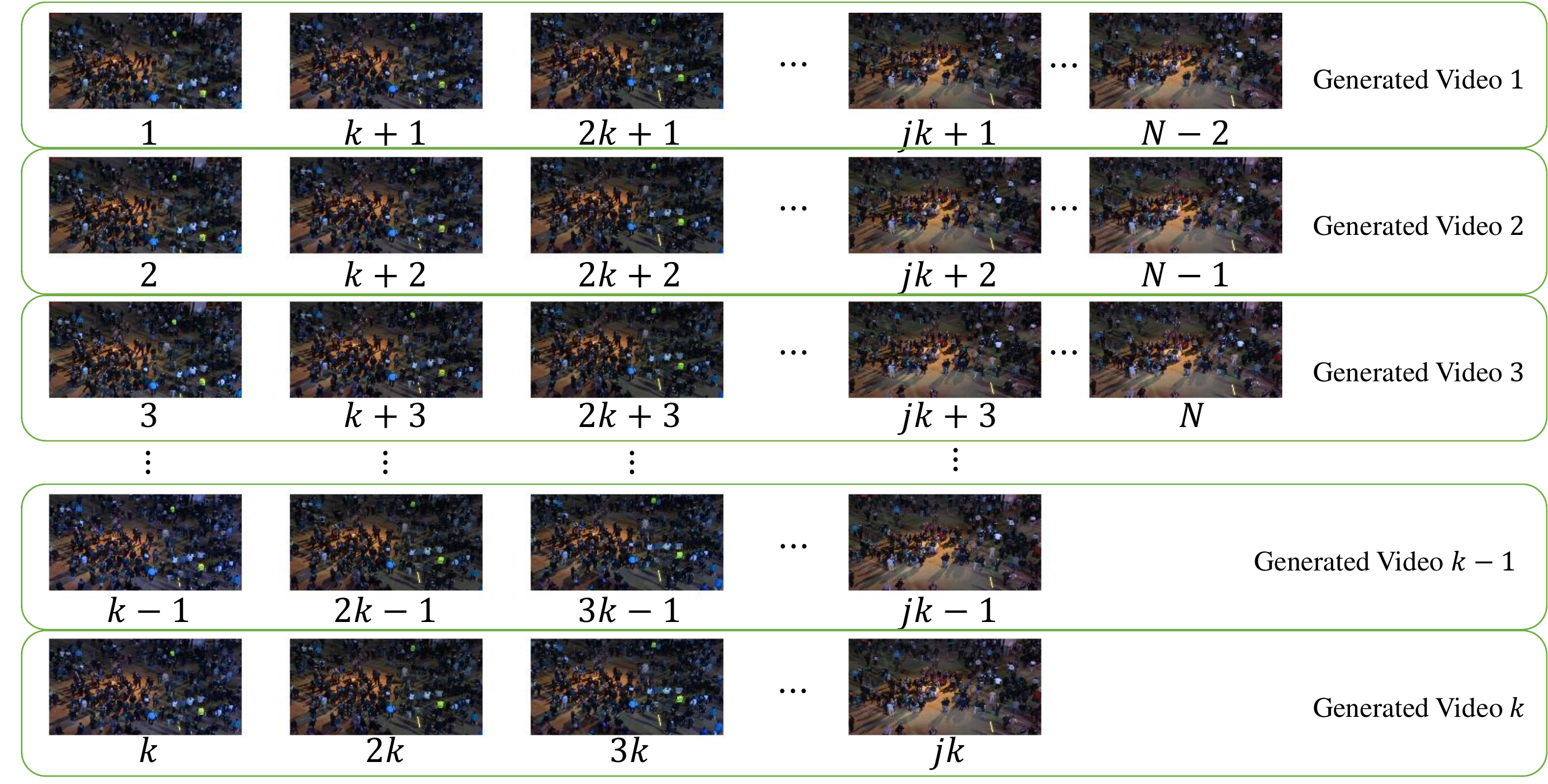}
    \caption{Illustration of multi-frame-rate dataset generation. We obtain $k$ new videos with $\frac{F}{k}$ fps given the original video with $F$ fps.}
    \label{fig:gen}
\end{figure}

For evaluating FraMOT, a multi-frame-rate dataset is necessary for training and evaluating the algorithm.
Instead of collecting and re-annotating extra data, we simulate the multi-frame-rate inputs from existing high-frame-rate MOT datasets. \wt{We do not consider a higher frame rate than 30 fps, because i) currently most cameras work at a frame rate of 30 fps, and streaming devices of higher frame rate are rare, and ii) to reduce computational cost most industrial scenarios tend to reduce the frame rate while not increase the frame rate,  and iii) higher frame rate usually does not bring extra challenge.} Given a video with $F$ frames per second and $N$ frames in total, the frames are denoted by $I_1, I_2, \ldots, I_N$. The target frame rate is $F' (F' < F)$ frame per second and we assume that $F$ is divisible by $F'$. Our simulation method re-samples the original video and re-composes the frames into new videos, satisfying the target frame rate. Specifically, we generate $
k=\frac{F}{F'}$ new videos, the $i$-th of which consists of frames $\hat{I}_{ij}, j=1,2,\ldots,\frac{N}{k}$ where 
\begin{equation}
    \hat{I}_{ij} = I_{(j-1)k+i}.    
    \label{eq:e1}
\end{equation}
We call $k$ the sampling factor of a simulation. As shown in Fig.~\ref{fig:gen}, the original video is decomposed into a matrix of $k$ rows and $\frac{N}{k}$ columns, with each row generating a new video at the target frame rate.  
%Using the simulation videos, we are then able to address the FraMOT problem either for training or for evaluation.
 \begin{table*}
 \small
 \centering
 \begin{tabular}{cc@{\ \ \ }c@{\ \ \ }c@{\ \ \ }c@{\ \ \ }c@{\ \ \ }c@{\ \ \ }c@{\ \ \ }c@{\ \ \ }c@{\ \ \ }c@{\ \ \ }c@{\ \ \ }c@{\ \ \ }c@{\ \ \ }c@{\ \ \ }c@{\ \ \ }c}
 \cline{1-17}
 \multirow{3}*{\centering Dataset} & \multicolumn{2}{c}{$k$=1} & \multicolumn{2}{c}{\centering $k$=2} & \multicolumn{2}{c}{\centering $k$=4} & \multicolumn{2}{c}{\centering $k$=8} & \multicolumn{2}{c}{$k$=16} & \multicolumn{2}{c}{\centering $k$=25} & \multicolumn{2}{c}{\centering $k$=36} & \multicolumn{2}{c}{\centering $k$=50} \\
 \cline{2-17}
 ~ & fr. & id. & fr. & id.& fr. & id.& fr. & id.& fr. & id.& fr. & id.& fr. & id.& fr. & id.  \\
 \cline{1-17}
 MOT17-train & 759 & 78& 380 & 78& 190 & 77& 95 & 76& 47 & 73& 30 & 71& 21 & 68& 15 & 64 \\ 
 MOT17-test & 846 & -& 423 & -& 211 & -& 106 & -& 53 & -& 34 & -& 23 & -& 17 & - \\ 
 MOT20-train & 2233 & 554& 1116 & 553& 558 & 552& 279 & 551& 140 & 547& 89 & 543& 62 & 539& 45 & 534 \\ 
 MOT20-test & 1120 & -& 560 & -& 280 & -& 140 & -& 70 & -& 45 & -& 31 & -& 22 & -\\ 
 \cline{1-17}
    \end{tabular}
    \caption{Statistics of the generated videos. `fr." and `id." represent the number of frames and identities per video, respectively. \bai{``-" denotes lacking of data since the testing dataset is hidden by the organizers to avoid misuse.}}
    \label{tab:gen_stats}
\end{table*}

%\textbf{The selection of target frame rates.} 
%The next problem is to decide a proper range of target frame rates, which is able to reflect the demand of industrial scenarios. Obviously we are not going to test the trackers on every single different frame rate,  so we hope to reflect a wider range of different frame rate settings using the least number of target frame rates.
%Therefore we propose to choose the original frame rates divided by the powers of 2 as the target frame rates, with adjustments in order to align some critical numbers. 
Considering that most videos nowadays have a default frame rate of 25 fps, we start from 25 fps and sample videos with 12.5 fps, 6.25 fps, 3.125 fps, 1.5625 fps, 1 fps, 0.722 fps, and 0.5 fps for a thorough evaluation of the trackers' performance. That means we choose 1, 2, 4, 8, 16, 25, 36, and 50 for $k$ in the Eq.~\ref{eq:e1}, leading to eight different settings in total. By using a broader range of non-geometric acceleration coherence, it is possible to obtain a more comprehensive representation of the overall impact while maintaining a small evaluation time cost.  Although it is possible to sample videos of frame rates lower than 0.5 fps, 
%given the fact that 0.5 fps setting has already brought a huge challenge, we believe a frame rate setting lower than 0.5 fps is not good for MOT application. Besides, 
a 0.25 fps setting will generate videos with less than 10 frames, leading to weak test cases. This range may also be extended in the future as the demands change. 

%\subsection{Evaluation Datasets and Metrics}
%We evaluate our proposed approaches on the MOT17 and MOT20 datasets with multi-frame-rate simulations mentioned in Section~\ref{sec:simulation}.
Table~\ref{tab:gen_stats} shows the statistics of the simulation datasets. Every single frame from the original dataset is perfectly included in the new dataset. Lower target frame rate setting results in a larger number of videos but a shorter length for each generated video. However, the average number of different identities in each video remains stable.

\subsubsection{Metrics}
Following the CLEAR MOT metrics and HOTA metrics, we define the corresponding mean metrics over multi-frame-rate settings to reflect the overall performance of trackers on videos with diverse frame rates. Specifically, we use mean-HOTA (mHOTA) and mean-MOTA (mMOTA) as the primary indicators of frame rate agnostic tracking performance, where 
\begin{equation}
    \mathrm{mHOTA} = \frac{1}{\mid F_{all}\mid}\sum_{F'\in F_{all}}{\mathrm{HOTA}_{F'}},
\end{equation}
\begin{equation}
    \mathrm{mMOTA} = \frac{1}{\mid F_{all}\mid}\sum_{F'\in F_{all}}{\mathrm{MOTA}_{F'}}.
\end{equation}
$F_{all}$ is the set of all target frame rates. We also provide mean-IDF1 (mIDF1) in the main results for reference. Similarily, $\mathrm{mIDF1} = \frac{1}{\mid F_{all}\mid}\sum_{F'\in F_{all}}{\mathrm{IDF1}_{F'}}$.

To further investigate the robustness of the trackers against multiple frame rates, we further propose a new metric named `Vulnerable Ratio" (VR):

\begin{equation}
    \mathrm{VR} = \frac{\mathrm{HOTA_{highest}} - \mathrm{HOTA_{lowest}}}{\mathrm{HOTA_{highest}}}, 
\end{equation}

\noindent where $\mathrm{HOTA_{highest}}$ and $\mathrm{HOTA_{lowest}}$ denote the highest HOTA and lowest HOTA among all different frame rate settings. VR indicates the largest possible drop in extreme cases and will be helpful in reliability and validity evaluation.

\subsection{Implementation Details}
\noindent\textbf{Joint extractor}.
We adopt YOLOX~\citep{DBLP:journals/corr/abs-2107-08430} detection framework with an extra identity branch as the joint extractor model. The backbone type is yolox-x. All baselines in the experiments share the same joint extractor architecture, including a detection branch and an identity branch (for tracking).

\noindent\textbf{Training data.} The joint extractor is first pre-trained on the COCO dataset without the identity branch and then trained on joint MOT17/20~\citep{MOT16,dendorfer2021motchallenge,dendorfer2020mot20}, CrowdHuman~\citep{shao2018crowdhuman}, CityScapes~\citep{Cordts2016Cityscapes} and HIE~\citep{lin2020human} datasets. For MOT datasets, we split the datasets into two different parts without overlapping with each other. The first part is for training, occupying about 60\% of the original datasets, and the second part is for evaluation and ablation studies. The training part is further divided into two sets, \ie, Set-A and Set-B, which share some frames but also have their own independent data. Set-A and other non-MOT datasets are for the joint extractor training, and Set-B is for joint extractor fine-tuning (smaller learning rate) and association module training. Set-B samples 300 images from each video, with 200 images shared with Set-A. \wt{The design of Set-A and Set-B can help avoid the association module from overfitting to the seen data.}

\noindent\textbf{Data augmentation.} We follow the same data augmentation techniques as YOLOX on non-tracking data (identity branch and association module will not be optimized on these data). On tracking data, we only apply random flip and random resize and remove other augmentations in YOLOX.

\noindent\textbf{Optimization.} Stochastic Gradient Descent (SGD) is used as our optimizer. We first train the joint extractor without association module on Set-A and other non-MOT datasets for 60 epochs. In this stage, we apply a cosine learning rate strategy with an initial learning rate of 0.00001. The batch size is set to 32. Then, we fine-tune the joint extractor and train the association module following the proposed Periodic Training Scheme (PTS) on Set-B for 10 more epochs each period. 
%In this stage, we apply step learning rate strategy with initial learning rates $\lambda_E=0.0000001$ and $\lambda_A=0.001$. 
The period number $N_p$ is set to 3.
% More details can be found in the appendix.

\noindent\textbf{Other hyper-parameters.} In the training stage, loss weights $\alpha$ is set to 0.5 and $\beta$ is set to 1. During online tracking, $\lambda_{high}$ is set to 0.6 and $\lambda_low$ is set to 0.1. The drop interval $\lambda_i$ is set to 30. A matching pair is formed only if the association score is greater than 0.1. 

 \begin{table*}[t]
 \footnotesize
 \centering
 \begin{tabular}{cc@{\ }c@{\ }c@{\ }c@{\ }c@{\ \ \ }c@{\ }c@{\ }c@{\ }c}
 \cline{1-10}
 \multirow{3}*{\centering Dataset} & \multirow{3}*{\centering Method} & \multicolumn{4}{c}{\centering known frame rate} & \multicolumn{4}{c}{\centering unknown frame rate}\\
 \cline{3-10}
 ~ & ~ & mHOTA$\uparrow$ & mMOTA$\uparrow$ & mIDF1$\uparrow$ & VR$\downarrow$ & mHOTA$\uparrow$ & mMOTA$\uparrow$ & mIDF1$\uparrow$ & VR$\downarrow$\\
 \cline{1-10}
 \multirow{6}*{\centering MOT17} & ByteTrack(\cite{zhang2021bytetrack}) & 52.5 & 64.4 & 61.3 & 38.3 & 51.0 & 62.9 & 59.3 & 43.9 \\
 ~ & CenterTrack(\cite{zhou2020tracking}) & - & - & - & - & 47.0 & 60.5 & 54.8 & 24.6 \\
 ~ & FairMOT(\cite{zhang2021fairmot}) & 51.0 & 60.1 & 56.9 & 32.3 & 49.7 & 58.0 & 56.0 & 35.5 \\
 ~ & GSDT(\cite{wang2021joint}) & 48.6 & 52.3 & 56.8 & 32.1 & 46.8 & 50.2 & 54.7 & 37.1 \\
 ~ & TraDeS(\cite{wu2021track}) & - & - & - & - & 38.6 & 58.3 & 44.4 & 49.0 \\
 ~ & Ours & \best{57.0} & \best{70.0} & \best{67.3} & \best{14.6} & \best{53.8} & \best{65.6} & \best{62.1} & \best{22.4} \\
 \cline{1-10}
 \multirow{4}*{\centering MOT20} & ByteTrack(\cite{zhang2021bytetrack}) & 43.5 & 48.6 & 49.9 & 58.1 & 40.7 & 46.6 & 46.9 & 63.3\\
 ~ & FairMOT(\cite{zhang2021fairmot}) & 42.2 & 36.6 & 48.4 & 49.9 & 41.0 & 34.2 & 47.1 & 53.1 \\
 ~ & GSDT(\cite{wang2021joint}) & 44.0 & 41.9 & 51.7 & 50.1 & 42.9 & 46.9 & 51.9 & 54.0 \\
 ~ & Ours & \best{50.8} & \best{59.3} & \best{58.3} & \best{38.8} & \best{48.0} & \best{56.1} & \best{53.5} & \best{41.0} \\
 \cline{1-10}
    \end{tabular}
    \caption{Overall results of recent state-of-the-art MOT methods and our method on MOT17 and MOT20, FraMOT versions. $\uparrow$ means higer is better, $\downarrow$ means lower is better. \best{Bold} for best results. (mHOTA, mMOTA, and mIDF1 are presented in the form of percentages.) Our method achieves the best performance under FraMOT settings both with known frame rate mode and unknown frame rate mode.}
    \label{tab:overall}
\end{table*}

\begin{figure*}[t]
    % \centering

    % \begin{subfigure}[b]{1.0\textwidth}
    % \centering
    % \includegraphics[width=6.1in]{curve_MOT17-crop.pdf}
    % \caption{MOT17 results on different input frame rate $F'$ ($F'=F/k$).}
    % \label{fig:my_label_s1}
    % \end{subfigure}
    
    % \begin{subfigure}[b]{1.0\textwidth}
    % \centering
    \includegraphics[width=6.25in]{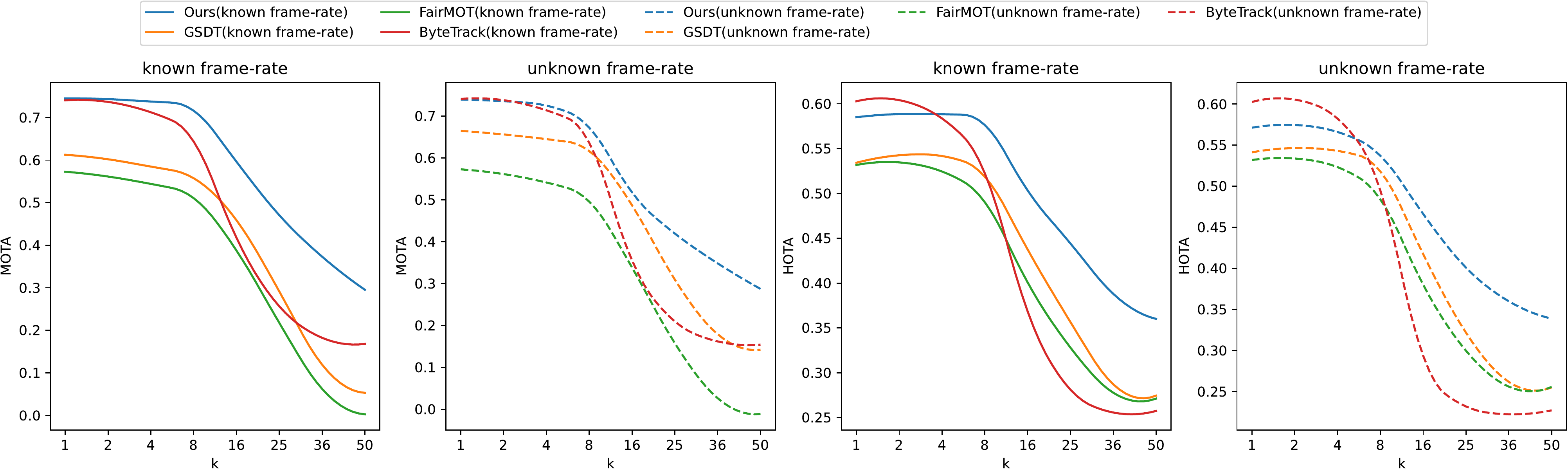}
    % \caption{MOT20 results on different input frame rate $F'$ ($F'=F/k$).}
    % \label{fig:my_label_s2}
    % \end{subfigure}

    \caption{MOTA and HOTA performance curves of recent state-of-the-art methods and our method on the MOT20 testing set (FraMOT versions) with both known frame rate and unknown frame rate.  
    %\bai{would be better to split the know and unknown into two sub-figures}
    }
    \label{fig:curve20}
\end{figure*}

 \begin{table*}[t]
 \small
 \centering
 \begin{tabular}{cc@{\ \ \ }c@{\ \ \ }c@{\ \ \ }c@{\ \ \ }c@{\ \ \ }c@{\ \ \ }c@{\ \ \ }c@{\ \ \ }c}
 \cline{1-10}
 \multirow{3}*{\centering Setting} & \multirow{3}*{\centering Param-Scale} & \multirow{3}*{\centering FAAM} & \multirow{3}*{\centering PTS} & \multicolumn{3}{c}{\centering known frame rate} & \multicolumn{3}{c}{\centering unknown frame rate}\\
 \cline{5-10}
 ~ & ~ & ~ & ~ & mHOTA$\uparrow$ & mMOTA$\uparrow$ & VR$\downarrow$ & mHOTA$\uparrow$ & mMOTA$\uparrow$ & VR$\downarrow$ \\
 \cline{1-10}
 Baseline (UM) & 1x &  &  & - & - & - & 56.5 & 74.6 & 34.6 \\
 Baseline (MM) & 3x &  &  & 58.9 & 77.0 & 29.5 & - & - & -\\
 %MM + PTS &  &  & \checkmark & 60.5 & 77.3 & 28.1 & - & - & -\\
 UM + PTS & 1x &  & \checkmark & - & - & - & 58.6 & 75.6 & 26.7 \\
 FAAM & 1x & \checkmark &  & 59.3 & 77.4 & 28.5 & 59.1 & 75.8 & 29.1 \\
 FAAM + PTS & 1x & \checkmark & \checkmark & 61.1 & 77.6 & 24.8 & 61.0 & 77.3 & 24.9 \\
 \cline{1-10}
    \end{tabular}
    \caption{Ablation study about the effectiveness of the proposed approaches.}
    \label{tab:ablation}
\end{table*}

\subsection{Results}

% MOT17 known framerate, MOT17 known framerate, MOT20 known framerate MOT20 unknown framerate (curve-HOTA, curve-MOTA, table-mMOTA, table-mHOTA, table-VR, table m-Recl, table m-Prec, table-IDSW, table-IDF1, table-time)
Table~\ref{tab:overall} shows the overall results of the recent open-source state-of-the-art MOT methods and our method on the challenging MOT17 and MOT20 datasets, with the aforementioned multiple frame rate simulation. ByteTrack~\citep{zhang2021bytetrack}, FairMOT~\citep{zhang2021fairmot}, and GSDT~\citep{wang2021joint} have design strategies using the provided frame rate to control cache length.
%so have different results for known frame rate mode. 
CenterTrack and TraDeS do not design any frame-rate-relevant procedures so their known frame rate mode results are absent.  Our method outperforms all other methods with all metrics in terms of frame rate agnostic tracking. In the known frame rate mode, we are 4.5\% and 6.8\% higher than the runner-up methods in MOT17 and MOT20 for mHOTA, respectively. For mMOTA, we are 5.6\% and 10.7\% higher. For mIDF1, we are 6.0\% and 6.6\% higher. Besides, we have the least Vulnerable Rate (VR), which means our performance is the least influenced by the change of frame rate.  In unknown frame rate mode, our method still outperforms other methods. The improvement in this mode is smaller than those of the known frame rate mode due to the difficulty of obtaining accurate motion distribution. We have 2.8\% and 5.1\% improvement in the term of mHOTA for MOT17 and MOT20, respectively. We also have higher mHOTA and mIDF1 scores in this mode, and the Vulnerable Rate is also lower than other methods. The results show that our proposed method is effective for frame rate agnostic tracking.
%Fig.~\ref{fig:curve17} and 

Fig.~\ref{fig:curve20} further shows the HOTA and MOTA curves of these methods w.r.t the sampling factor $k$ (larger $k$ means lower frame rate) on MOT20 FraMOT simulation datasets.
Among these methods, ByteTrack is the first method making use of the YOLO-X detection baseline, and thus has more accurate detection results and outperforms all other trackers in normal frame rate setting ($k=1$). To compare with ByteTrack fairly, we also develop our method based on YOLO-X baseline. However, ByteTrack does not use any appearance features for tracking, while our method has an extra tracking branch for exploiting appearance features. As can be observed in Fig.~\ref{fig:curve20}, our method is slightly lower than ByteTrack at $k\leq 4$ in terms of HOTA, which confirms that appearance features do not help improve and may harm the tracking performance with high frame rates. However, when the frame rates are low (\eg, with larger $k$), the performance of ByteTrack drops significantly \bai{due to less reliable motion}. At the same time, the performance of our method also drops slowly but is much higher than all compared methods. \bai{It shows that the tracking becomes more difficult when the frame rate is lowered, but our method has better capability to handle these complicated scenarios}. Nevertheless, the overall performance of our method is the best, which reveals our design of FAAM and PTS training is effective for FraMOT.

Other than the MOTChallenge benchmarks, we also conducted experiments using the recent SOMPT22 dataset~\citep{simsek2023sompt22}, and the results are presented in the applendix~\ref{secA2}.
%\bai{which reveals that appropriately using both motion and appearance features is helpful for FraMOT.} 

%which we find is necessary for frame rate agnostic tracking. According to the authors of the ByteTrack method, they found that integrating a Re-ID branch into the YOLO-X detection baseline did not improve the tracking performance. In our experiments, we also find that directly adding a Re-ID branch in the YOLO-X will have bad influence on detection accuracy, and thus harm the overall tracking performance. Due to this reason, our YOLO-X baseline with an extra tracking branch is not performing as good as the original ByteTrack model in terms of detection accuracy, and thus our method is slightly lower than ByteTrack at $k\leq 4$ in terms of HOTA. However, when the frame rate goes down, all methods can hardly handle the change and the performance drop dramatically. In this challenging scenario, our method handle the change much better and thus has the most steady performance. \bai{which reveals XXX}

\subsection{Ablation Study and Analysis}
To better understand the proposed methods, we further conduct quantitative experiments on the MOT20 validation set (which is split from the original training set and is not included in the real training set). 
\subsubsection{Effects of the Proposed Methods}
Table~\ref{tab:ablation} shows how different components affect the overall performance. In this experiment, we first present two straightforward baselines for the FraMOT problem, \ie, Unified Model (UM) and Multiple Model (MM). Baseline (UM) is a simple model which is directly trained from sampled frame pairs of all different frame rates without any extra strategy proposed in this work. 
In contrast, Baseline (MM) trains 3 separate models for 3 different ranges of sampling factor $k$, \ie, high-frame-rate ($k < 6$), middle-frame-rate ($ 6 \leq k < 20$) and low-frame-rate ($20 \leq k$), and deploys each frame rate specific model independently for the corresponding frame rate during testing.
%is a multi-model baseline with 9 models. When training and testing these models, we feed different frame rate samples to different models, each of them has its own working range of frame rates. 
The first baseline (UM) does not need a frame rate number during inference and thus is presented in the unknown frame rate mode. The second baseline (MM) requires a frame rate number for model switching, so its results are presented in known frame rate mode. \wt{All methods in this ablation study use the same joint extractor framework and checkpoint, with differences in the association modules.} The mean-HOTA, mean-MOTA, and Vulnerable Ratio of the UM baseline are the worst among all settings in the table, indicating traditional unified association models are not capable of performing frame rate agnostic tracking. The MM baseline is better than the unified model baseline due to the prior knowledge of frame rate division. However, the results are still not satisfying and the method is not feasible when the frame rate number is not given, which is not comparable with the way that humans track objects. 
Our proposed methods are presented by `UM + PTS', `FAAM', and `FAAM + PTS'. `UM + PTS' means applying the Periodic Training Scheme on the unified model baseline. With the help of the PTS strategy, the unified model baseline improves by 2.1\% mHOTA and 1.0\% mMOTA, as well as reduces the Vulnerable Ratio by 8.9\%, showing that the PTS training successfully reduces the difficulty of frame rate agnostic tracking training. `FAAM' means only applying the Frame Rate Agnostic Association Module, which can both work at known frame rate mode and unknown frame rate mode. This unified approach performs better than both baselines while using only 1/3 of the parameters compared with the MM baseline. More importantly, it can automatically infer the frame rate information.  
%We also apply PTS to the multi-model baseline and reduce the number of models to 4. With the help of PTS strategy, the multi-model baseline improves by 0.9 mHOTA with less models. 
%These two experiments shows that the PTS training successfully reduce the difficulty of frame rate agnostic tracking training. 
When we apply both the two proposed methods, presented as `FAAM + PTS', we obtain the best results of 61.1\% mHOTA and 77.6\% mMOTA, with the lowest Vulnerable Ratio of 24.9\% (known frame rate mode). The results in the known frame rate mode are also improved and are quite competitive compared with the known frame rate mode. The results above show that both two approaches are effective, improve the overall frame rate agnostic tracking accuracy and make the tracker wiser.
% metrics mHOTA, mMOTA
% baseline (single-frame-rate trained + handcrafted)
% % baseline (multi-frame-rate trained + handcrafted)
% baseline (single-frame-rate trained + assoc)
% baseline (multi-frame-rate trained + assoc)
% ours (multi-frame-rate trained + FAA)
% ours (multi-frame-rate trained + PTS)
% ours (multi-frame-rate trained + PTS + FAA)
% ours (multi-frame-rate trained + PTS + FAA - appr)

\subsubsection{Results on Unseen Frame Rates}

We have selected 8 different sampling factors $k$ in the training and testing dataset simulation. To better understand the behaviors of the proposed method about handling various frame rates, we conducted experiments on unseen frame rates, \ie, testing on frame rates that are different from the training set. \wt{Specifically, we select $k=1,3,6,12,21,30,43,62$ and rebuild the validation set based on these unseen $k$. We still include the normal frame rate,  \ie, $k=1$, in the unseen set because it is a baseline for analysis.} Table~\ref{tab:unseen} shows the results of two baselines and our FAAM equipped with PTS training. All models drop slightly when tested on unseen frame rates. However, our FAAM still outperforms the two baselines. Note that the FAAM result here is from unknown-frame-rate mode, which does not need extra frame rate information, while the MM baseline must be provided with a frame rate number.

 \begin{table}
 \small
 \centering
 \begin{tabular}{cc@{\ \ \ }c@{\ \ \ }c@{\ \ \ }c@{\ \ \ }c@{\ \ \ }c@{\ \ \ }c@{\ \ \ }c@{\ \ \ }c}
 \cline{1-4}
 \multirow{3}*{\centering Setting} & \multicolumn{3}{c}{\centering unseen frame rate}\\
 \cline{5-10}
 ~ & mHOTA$\uparrow$ & mMOTA$\uparrow$ & VR$\downarrow$ \\
 \cline{1-10}
 Baseline (UM) & 54.9 & 72.1 & 36.7 \\
 Baseline (MM) & 57.0 & 74.4 & 33.5 \\
 %MM/4 + PTS & 59.3 & 75.4 & 31.7 \\
 FAAM + PTS & 59.6 & 75.0 & 29.6 \\
 \cline{1-4}
    \end{tabular}
    \caption{Performance on unseen frame rates.}
    \label{tab:unseen}
\end{table}

\subsubsection{Results under Dynamic Sampling Setting}

\begin{figure}
    \centering
    \includegraphics[width=3.1in]{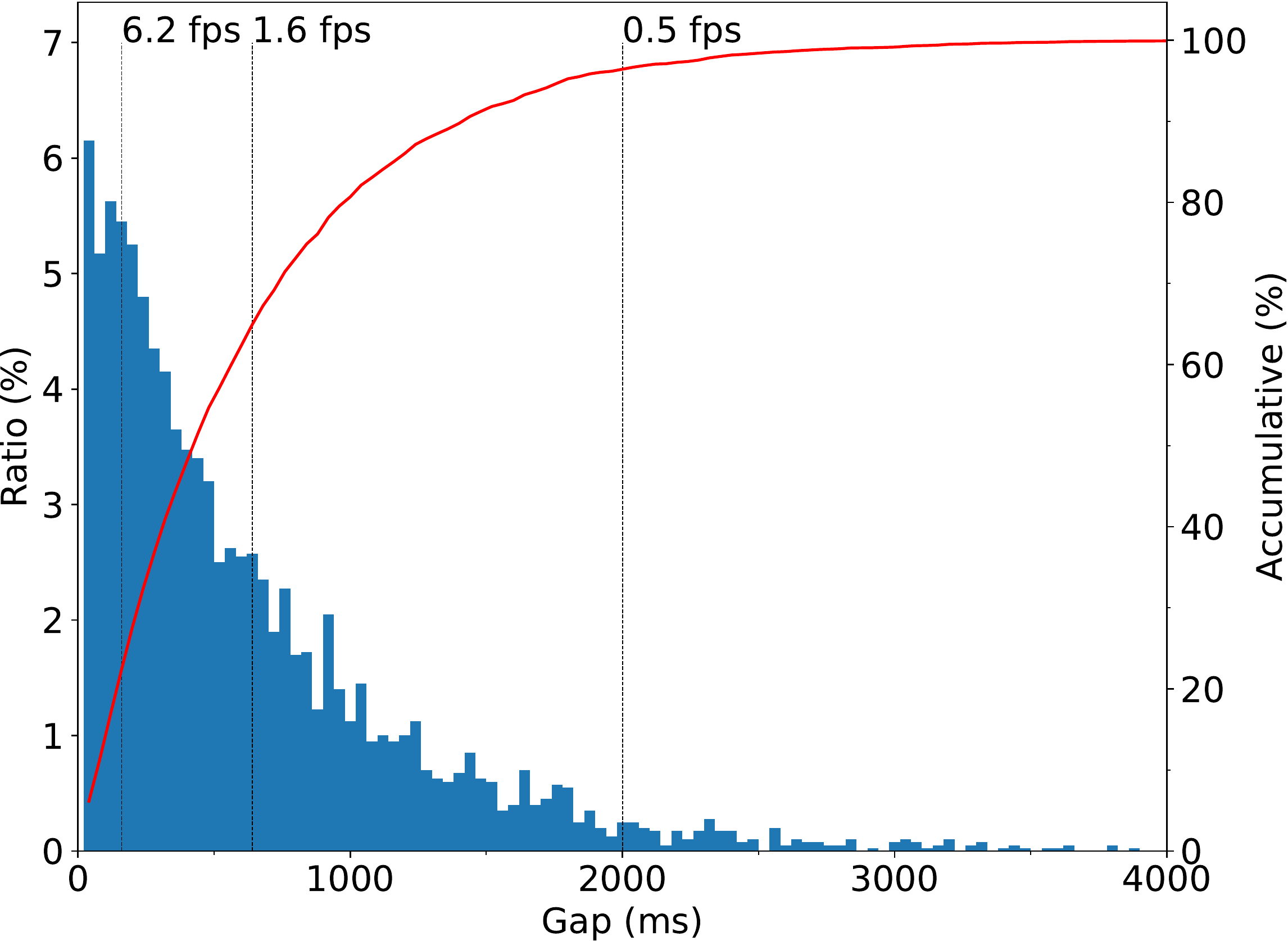}
    \caption{Sampling gap distribution of the generated videos under dynamic sampling setting.}
    \label{fig:gap}
\end{figure}

We also tried testing the methods on dynamic sampling videos to simulate some scenarios with network instability, where some images are accidentally, randomly missing. We generated dynamic sampling videos by randomly sampling frames in the original videos. Given a video with $n$ frames and an integer $k$, we randomly choose $\frac{n}{k}$ frames from the frame pool (initially includes all frames of the given video) and ensemble the chosen frames into a video. The chosen frames are then removed from the frame pool. We keep choosing frames randomly from the frame pool and generating videos until the frame pool is empty. In such way, we will obtain $k$ videos in total with approximately $\frac{n}{k}$ frames in each generated video. We chose $k=16$. Fig.~\ref{fig:gap} shows the distribution of the sampling gap, with an average gap time of 15.7 frames (628 ms) and a median gap time of 11 frames (440 ms).

Table~\ref{tab:dynamic} shows the results of two baselines and our FAAM equipped with PTS training. Under dynamic sampling settings, our FAAM still outperforms the two baselines. Note that the FAAM result here is from unknown-frame-rate mode, and the baseline (MM) treats all dynamic videos as 1.6 fps (as the average gap time is 628 ms, which is approximately 1.6 fps).

 \begin{table}
 \small
 \centering
 \begin{tabular}{cc@{\ \ \ }c@{\ \ \ }c@{\ \ \ }c@{\ \ \ }c@{\ \ \ }c@{\ \ \ }c@{\ \ \ }c@{\ \ \ }c}
 \cline{1-4}
 \multirow{3}*{\centering Setting} & \multicolumn{3}{c}{\centering dynamic sampling}\\
 \cline{5-10}
 ~ & HOTA$\uparrow$ & MOTA$\uparrow$ & IDF1$\uparrow$ \\
 \cline{1-10}
 Baseline (UM) & 50.5 & 76.2 & 58.3 \\
 Baseline (MM) & 47.6 & 74.4 & 53.3 \\
 %MM/4 + PTS & 59.3 & 75.4 & 31.7 \\
 FAAM + PTS & 55.5 & 77.5 & 64.8 \\
 \cline{1-4}
    \end{tabular}
    % \vspace{1mm}
    \caption{Performance under dynamic sampling setting.}
    \label{tab:dynamic}
\end{table}

\subsubsection{Influence of Different Motions}

\begin{table*}[]
\begin{tabular}{clllllllll}
\hline
\multirow{2}{*}{Model} & \multicolumn{3}{c}{Slow ($<$ 4 ppf)}                                         & \multicolumn{3}{c}{Normal (8-14 ppf)}                                 & \multicolumn{3}{c}{Fast (19 ppf)}                                            \\
                       & \multicolumn{1}{c}{mHOTA} & \multicolumn{1}{c}{mMOTA} & \multicolumn{1}{c}{VR} & \multicolumn{1}{c}{mHOTA} & \multicolumn{1}{c}{mMOTA} & \multicolumn{1}{c}{VR} & \multicolumn{1}{c}{mHOTA} & \multicolumn{1}{c}{mMOTA} & \multicolumn{1}{c}{VR} \\ \hline
Baseline (UM)          & 68.0                      & 72.2                      & 5.7                    & 50.7                      & 58.3                      & 14.3                   & 54.2                      & 52.4                      & 19.1                   \\
Baseline (MM)          & 68.7                      & 73.3                      & 5.1                    & 51.5                      & 58.9                      & 18.5                   & 44.1                      & 38.5                      & 51.9                   \\
FAAM + PTS             & 68.9                      & 72.3                      & 8.2                    & 54.5                      & 59.7                      & 7.0                    & 53.8                      & 50.7                      & 19.0                   \\ \hline
\end{tabular}
\caption{Performance on sub-datasets with different object motions.}
\label{tab:motion_expr}
\end{table*}

In order to investigate the effectiveness of our proposed method in different scenarios with varying object motions, we conducted additional experiments. To do this, we utilized the MOT17 validation set and divided it into three separate subsets based on the object motions present in each video sequence. To estimate the object motions, we computed the median motion per frame of the top 30\% fastest objects (in pixels per frame, ppf), which provided a clear representation of the motion differences between the videos.

The first subset, which consisted of sequences MOT17-02 and MOT17-04, featured static camera views and slow object motions (with a median motion of less than 4 ppf for the top 30\% fastest objects). The second subset included MOT17-05, MOT17-09, MOT17-10, and MOT17-11, which featured moving camera views and normal object motions (with a median motion of 8-14 ppf for the top 30\% fastest objects). The third subset comprised MOT17-13, which featured a camera view that was turning significantly and fast object motions (with a median motion of 19 ppf for the top 30\% fastest objects).

We evaluated two baseline methods and our proposed approach on each of these subsets and recorded the results in Table \ref{tab:motion_expr}. Our findings demonstrated that the baseline method MM performed better under slow scenarios but was ineffective under fast object motions, while the baseline method UM performed better under fast object motions but struggled under slow object motions. In contrast, our proposed approach maintained a balanced performance across all three scenarios and achieved the highest overall performance.

 \begin{table*}[t]
 \small
 \centering
 \begin{tabular}{ccccc@{\ \ \ }c@{\ \ \ }c@{\ \ \ }c@{\ \ \ }c@{\ \ \ }c}
 \cline{1-10}
 \multirow{3}*{\centering Group} & \multirow{3}*{$N_p$} & \multirow{3}*{threshold} & \multirow{3}*{criterion} & \multirow{3}*{mHOTA$\uparrow$} & \multirow{3}*{mMOTA$\uparrow$} & \multirow{3}*{VR$\downarrow$} & \multicolumn{3}{c}{\centering HOTA@k$\uparrow$}\\
 \cline{8-10}
 ~ & ~ & ~ & ~ & ~ & ~ & ~ & $k=1$ & $k=16$ & $k=50$\\ 
 \cline{1-10}
 \multirow{5}*{\centering Different $N_p$} & 0 & 0.9 & dist. & 59.1 & 75.8 & 29.1 & 65.5 & 60.0 & 46.4 \\
 ~& 1 & 0.9 & dist. & 60.6 & \textbf{77.4} & 26.9 & \textbf{65.6} & 62.3 & 48.3 \\
 ~& 2 & 0.9 & dist. & 60.8 & 77.2 & 25.8 & 65.2 & 62.6 & 48.8 \\
 ~& 3 & 0.9 & dist. & \textbf{61.0} & 77.3 & \textbf{24.9} & 65.3 & \textbf{62.7} & \textbf{49.4} \\
 ~& 4 & 0.9 & dist. & 60.7 & 77.2 & 27.2 & 65.3 & \textbf{62.7} & 47.9 \\
 \cline{1-10}
 \multirow{5}*{\centering Different thresholds} & 3 & 0.5 & dist. & 60.8 & \textbf{77.4} & 25.2 & 65.2 & 62.6 & 49.1 \\
 ~& 3 & 0.7 & dist. & 60.9 & 77.3 & 25.0 & 65.3 & 62.6 & 49.3 \\
 ~& 3 & 0.9 & dist. & 61.0 & 77.3 & 24.9 & 65.3 & 62.7 & 49.4 \\
 ~& 3 & 1.1 & dist. & 61.0 & 77.2 & 24.8 & 65.3 & 62.7 & 49.5 \\
 ~& 3 & 1.3 & dist. & \textbf{61.1} & 77.1 & \textbf{24.7} & \textbf{65.4} & \textbf{62.8} & \textbf{49.6} \\
 \cline{1-10}
 \multirow{3}*{\centering Different criteria} & 3 & 0.9 & random & 58.7 & 75.9 & 26.6 & 63.1 & 61.0 & 46.7 \\
 ~& 3 & 0.9 & sim. & 60.8 & \best{77.5} & 26.8 & \best{65.7} & \best{62.7} & 48.3 \\
 ~& 3 & 0.9 & dist. & \best{61.0} & 77.3 & \best{24.9} & 65.3 & \best{62.7} & \best{49.4} \\
 \cline{1-10}
    \end{tabular}
    \caption{Ablation studies about different hyper-parameters including period number $N_p$, matching threshold, and IBDV matching criteria.}
    \label{tab:ablation2}
\end{table*}

\begin{figure*}
    \centering
    
    \begin{subfigure}[b]{0.56\textwidth}
    \centering
    \includegraphics[width=3.6in]{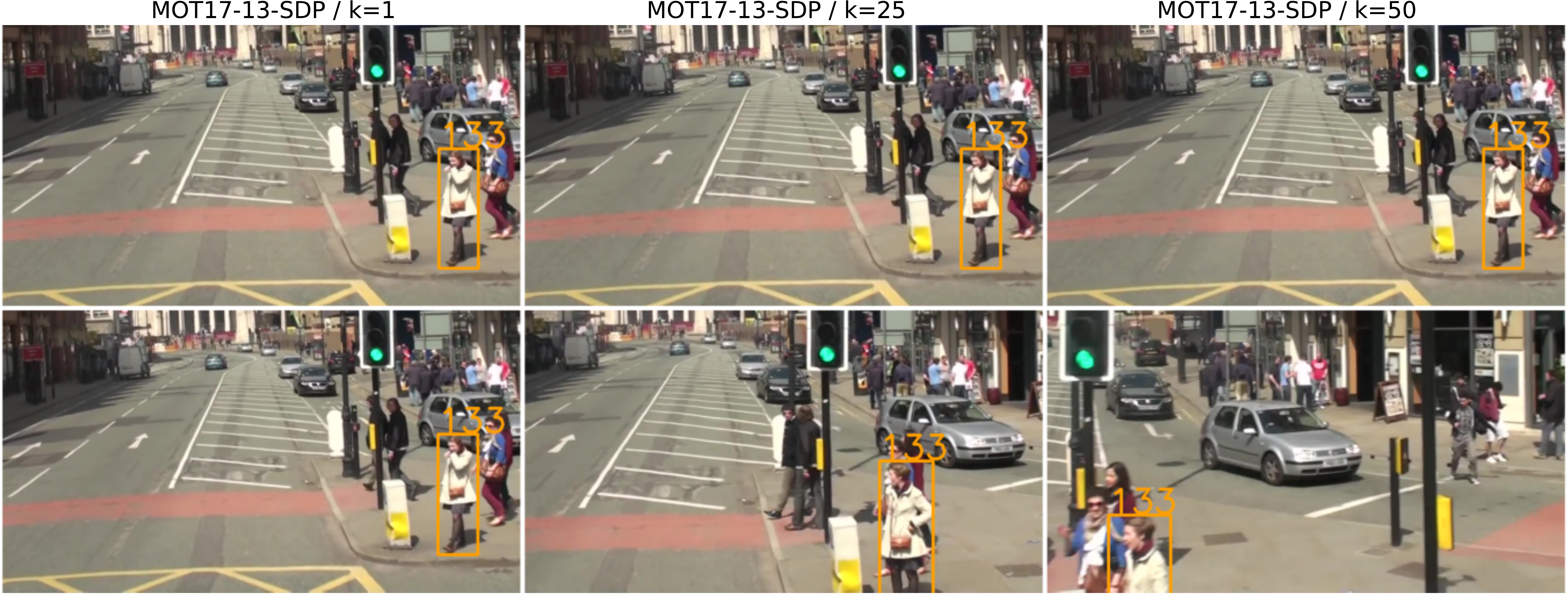}
    \caption{Sequence MOT17-13 with sampling factor $k=1,25,50$. ID 133 is emphasized.}
    \end{subfigure}
    % \hspace{1mm}
    \begin{subfigure}[b]{0.42\textwidth}
    % \centering
    \includegraphics[width=2.7in]{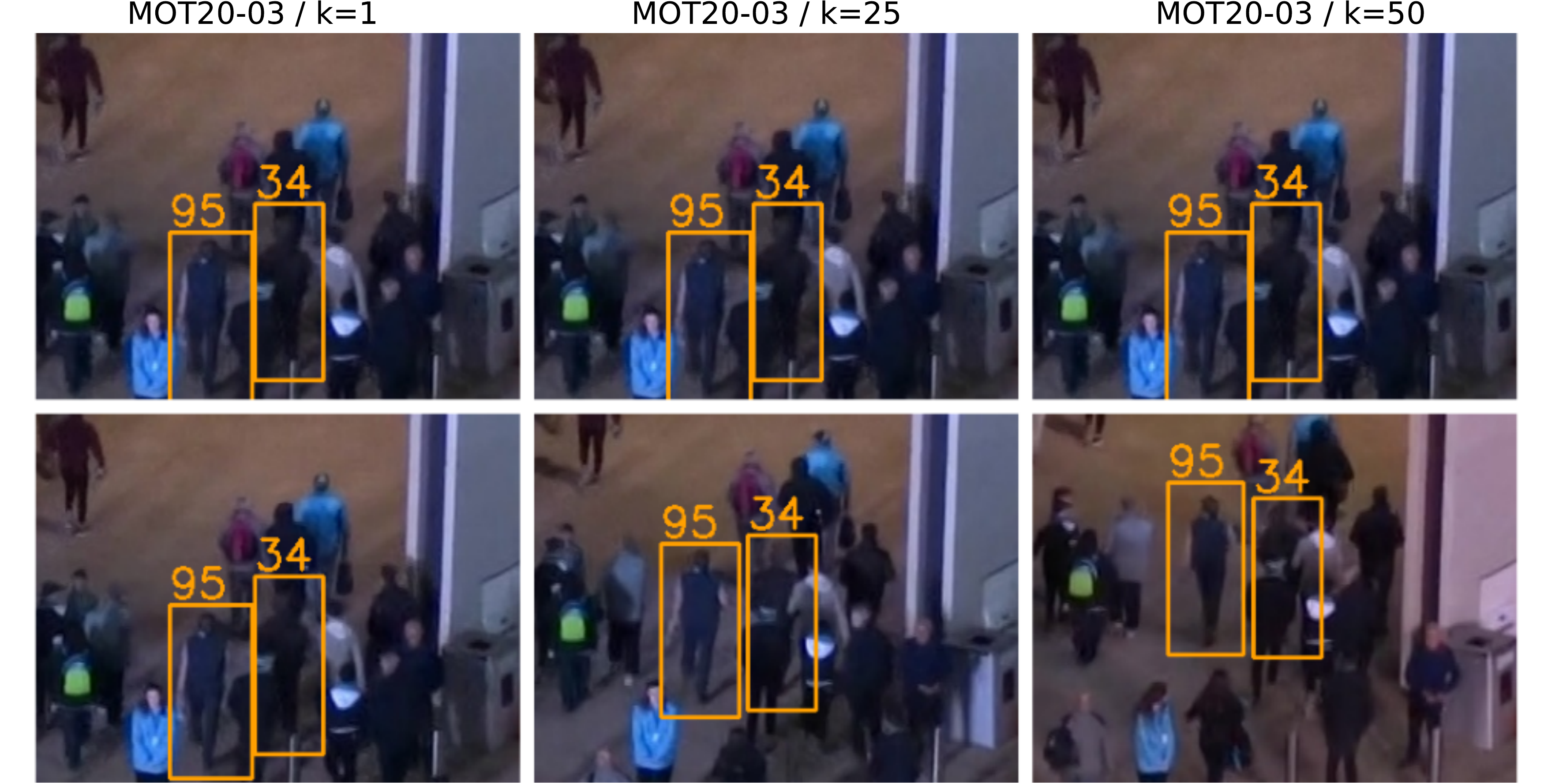}
    \caption{Sequence MOT20-03 with sampling factor $k=1,25,50$. ID 34 and 95 are emphasized.}
    \end{subfigure}

    \caption{Visualization of the multi-frame-rate simulation videos. The first row and the second row are adjacent frames. When the sampling factor $k$ is increased, the task becomes more challenging.}
    \label{fig:vis_img}
\end{figure*}

\begin{figure}
    \centering
    
    \begin{subfigure}[b]{0.5\textwidth}
    \centering
    \includegraphics[width=3.1in]{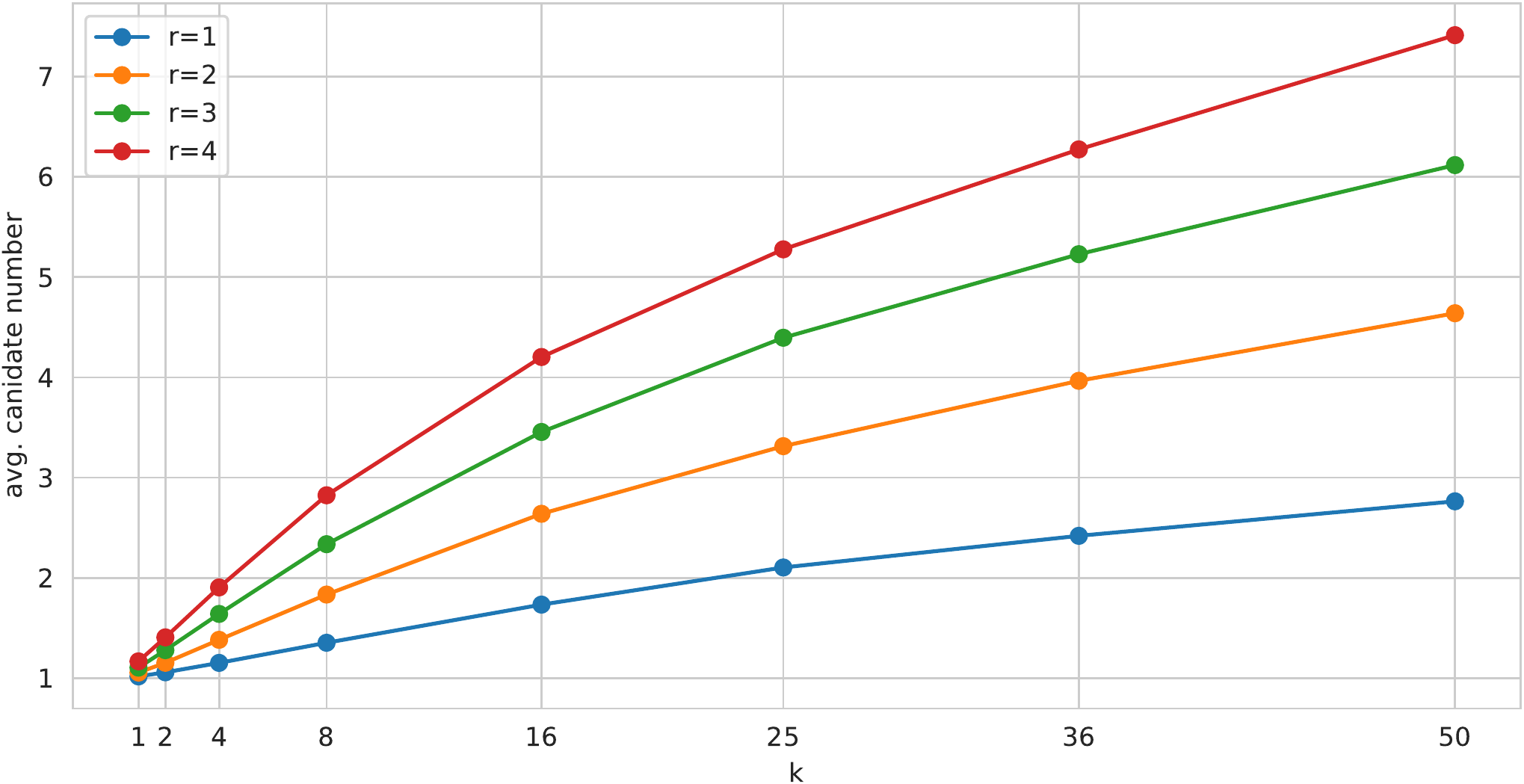}
    \caption{MOT17 training set.}
    \end{subfigure}
    % \hspace{1mm}
    \begin{subfigure}[b]{0.5\textwidth}
    % \centering
    \includegraphics[width=3.1in]{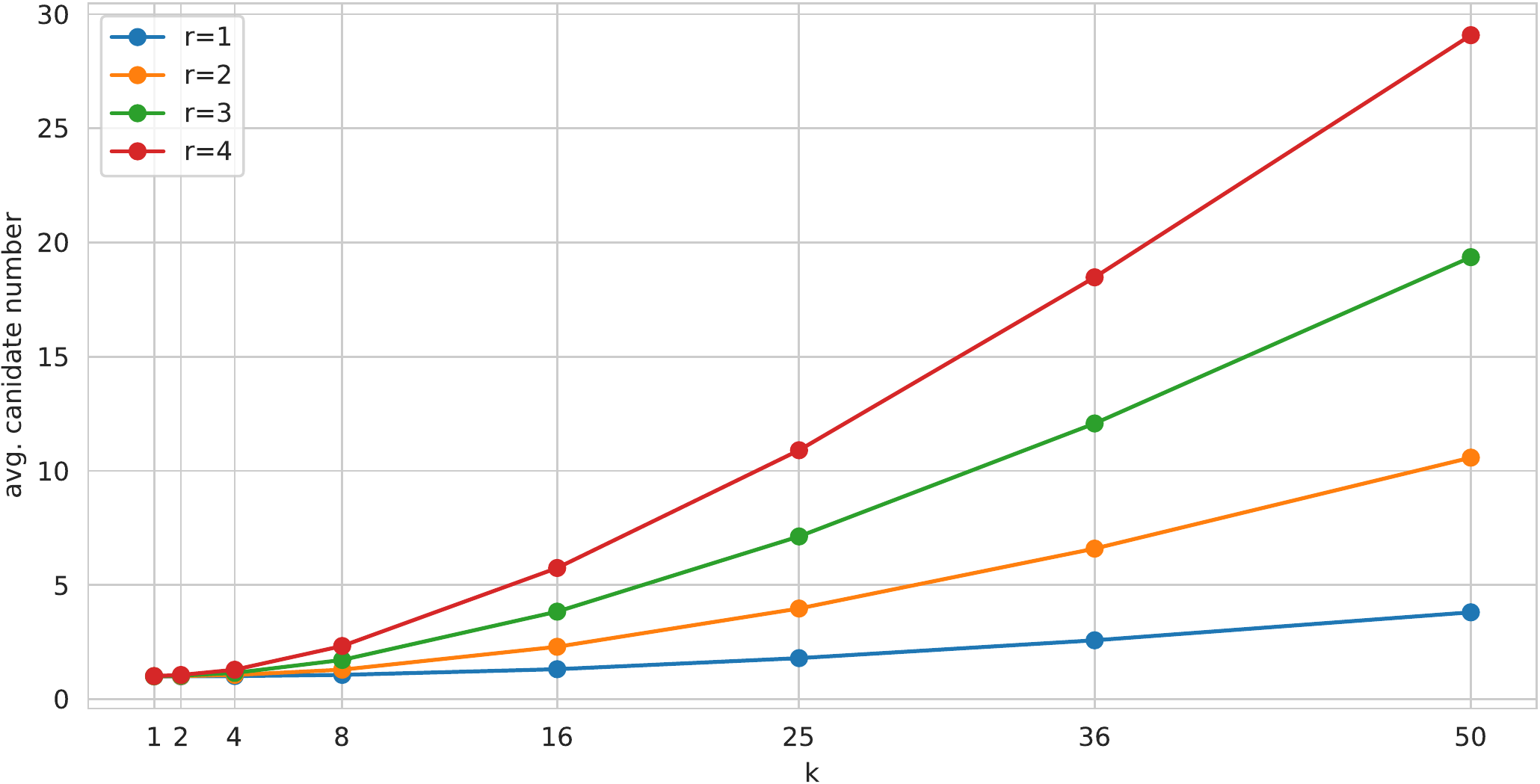}
    \caption{MOT20 training set.}
    \end{subfigure}

    \caption{Curves of candidate numbers w.r.t sampling factor $k$ and thresholding factor $r$ in MOT17 and MOT20 training set.}
    \label{fig:vis_can}
\end{figure}

\subsubsection{Influence of Hyper-Settings}
Table~\ref{tab:ablation2} shows how the hyper-settings influence the overall performance.

\noindent\textbf{Period number $N_p$.}
In PTS training, we introduce the period number $N_p$. To understand its influence, we conduct experiments with different $N_p$ from 0 to 4. 0 means PTS is not applied. We can see that when $N_p=3$, we obtain the best performance, while the other $N_p$ settings are also obtaining better performance than the not applied case. The results indicate that a large $N_p$ is not bringing more gain. It also suggests different checkpoints at later periods do not change the inference environment a lot, the critical change is at the first period.

% \noindent\textbf{Model size.}
\noindent\textbf{Matching threshold.}
Thresholding is sometimes affecting the results a lot. To ensure the matching threshold is stable and the gain is not from the thresholding strategy, we conduct experiments with different thresholds ranging from 0.5 to 1.3. We can see that our method obtains the best mHOTA at the threshold of 1.3, and obtains the best mMOTA at the threshold of 0.5. Although the thresholds vary, the performances are still stable with less than 0.3\% difference. We choose 0.9 as the final threshold because we can obtain a balance between mHOTA and mMOTA at the threshold of 0.9.

\noindent\textbf{IBDV criteria.}
In Section~\ref{sec:ibdv} we introduce Inter-frame Best-matched Distance Vector (IBDV) as the feature of frame rate embedding. We further design more different matching rules to understand how the rules affect the overall results. We present three different matching rules in the table. `random' means we randomly sample pairs as matched pairs. `sim.' means we choose the pairs with the highest appearance similarity as the matched pairs. `dist.' means we choose the pairs with closest positional distance as the matched pairs. The results show that both `sim.' and `dist.' rules perform better than the `random' rule, while the `dist.' rule is slightly better than the `sim.' rule. It might be because positional information is more reliable in general. It also reminds us that more complicated rules might be more effective. \label{sec:bestmatching}

\subsubsection{Visualization and Analysis}
In this section, we demonstrate the data simulation and our proposed method via visualization and analysis.

\noindent\textbf{Visualization of FraMOT simulation data.}
Fig.~\ref{fig:vis_img} shows some selected simulation image data from sequence MOT17-13 and sequence MOT20-03. In MOT17-13, the concerned target (ID 133) has a small movement between adjacent frames when $k=1$, while the movement becomes much larger at $k=50$. In MOT20-03, the movements (ID 34 and 95) are moderated, but the numbers of matching candidates are still increased a lot in this crowded scenario.

Fig.~\ref{fig:vis_can} illustrates curves of candidate numbers during matching with respect to sampling factor $k$ and thresholding factor $r$, \ie, how many possible candidates that are having an equal or closer distance to the concerned object, compared with the $r$ times distance to the correct ground-truth object. In other words, the curves show the average numbers of objects we must compare with during matching, under different $k$ and different thresholding strategies. Usually, $r$ is set to a number larger than 1 in order to make the correct object with the same identity number being compared for sure, or we are not able to recall the previously tracked object. If the candidate number is large, that means we have more chances to make mistakes and thus the task is more challenging. From the two figures we can see that when $k$ is increased, the candidate numbers increase significantly, especially in MOT20 dataset. In fact, in normal frame rate scenarios, the candidate number is slightly larger than 1, which means in most cases, finding the object with the closest distance will hit the ground-truth. It also well explained why introducing Re-ID branch in YOLO-X baseline did not improve the results at the normal frame rate,  it is because positional information is so reliable to perform matching. Fig.~\ref{fig:vis_can} shows that the FraMOT task is much more difficult due to various candidate numbers. 

\begin{figure}
    \centering
    \includegraphics[width=3.0in]{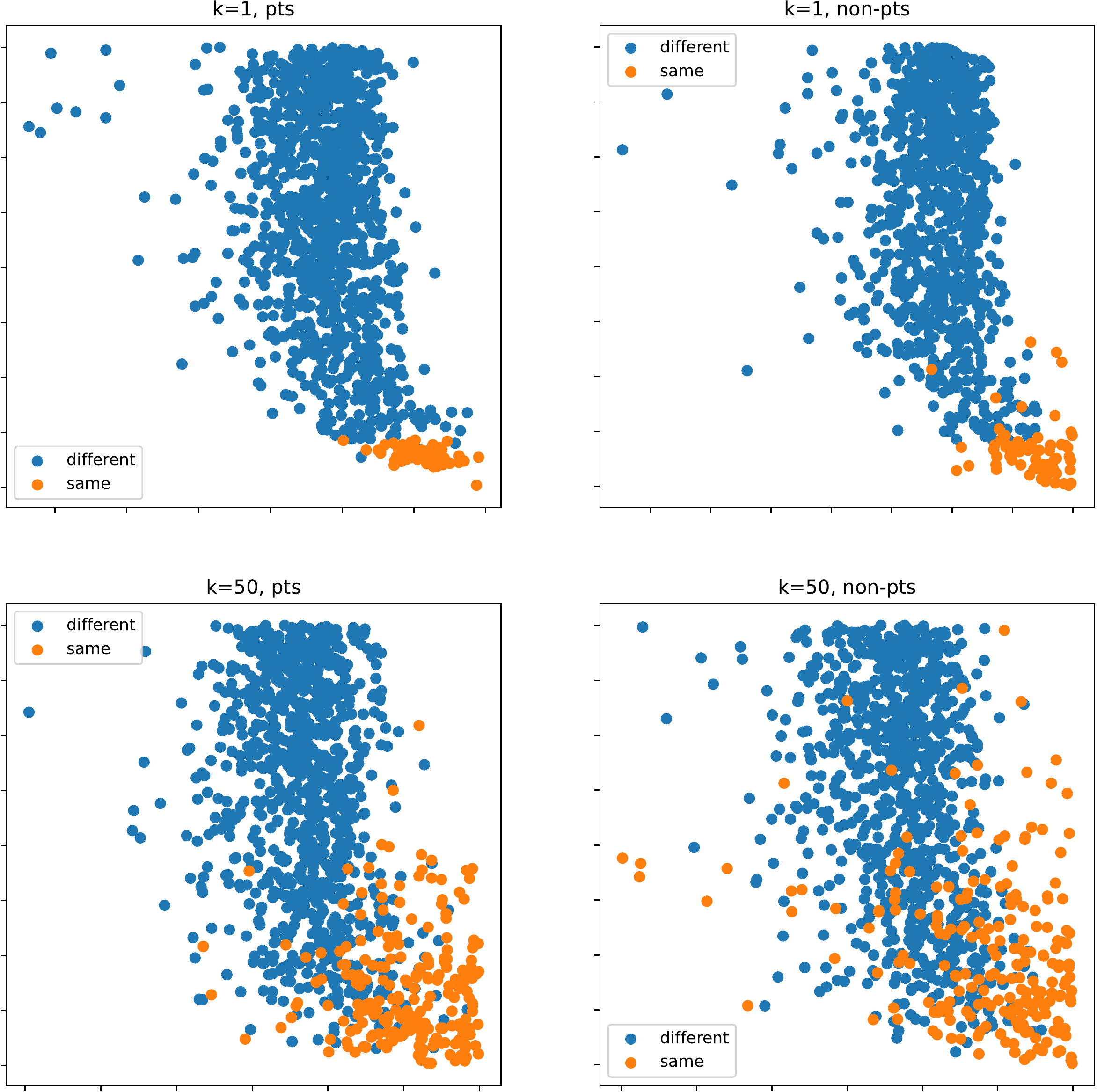}
    \caption{Visualization of affinity features w.r.t positional cues and appearance cues. Applying the PTS strategy leads to a clearer boundary between the same and different identity pairs.}
    \label{fig:affeat_vis}
\end{figure}

\noindent\textbf{Visualization of the PTS data.}
\ref{fig:affeat_vis} visualizes the affinity features used in this paper. We both show PTS and non-PTS affinity features at sampling rates of $k=1$ and $k=50$. The four sub-figures show that after applying PTS training, the boundary between the same and different identity pairs is clearer, which means the task is easier. At both sampling rates, there are more difficult data points in non-PTS counterparts. We can further find that at $k=50$, the gap between PTS and non-PTS is enlarged compared with $k=1$, which indicates PTS is more effective at larger sampling rates.  

\begin{figure}
    \centering
    \includegraphics[width=2.8in]{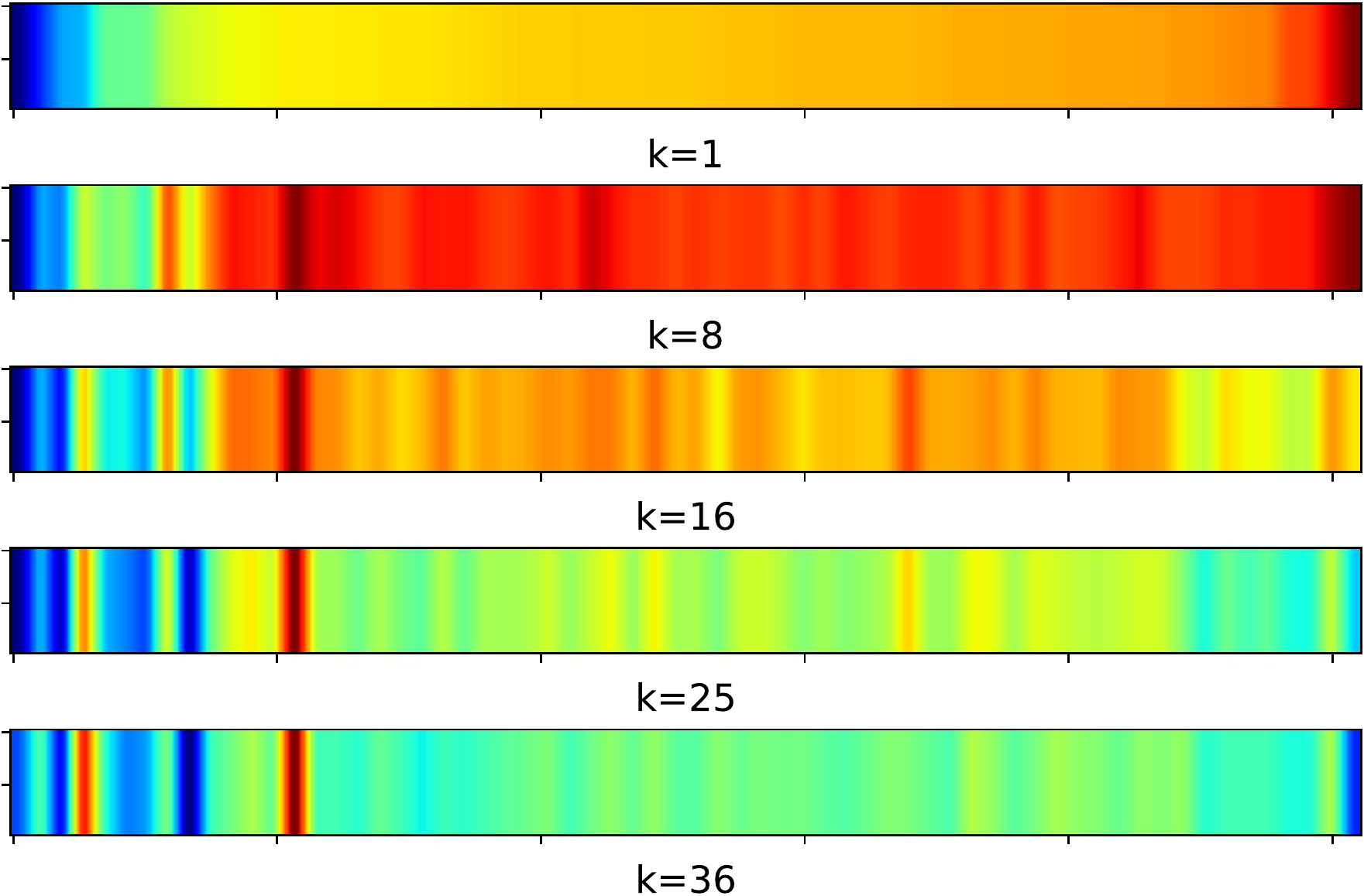}
    \caption{Attention map of the Frame Rate Agnostic Association Module (FAAM).}
    \label{fig:atten_vis}
\end{figure}

\noindent\textbf{Visualization of the attention embedding of Frame Rate Agnostic Association Module.}
\ref{fig:atten_vis} illustrates the attention maps learned by the proposed Frame Rate Agnostic Association Module (FAAM) at different sampling rates (different $k$-s). The channels of the attention map at $k=1$ are sorted in increasing order. Other attention maps are aligned with the $k=1$ attention map. When $k$ is small, \eg, $k<16$, the right part of the affinity features are emphasized. Then, the right part becomes less important when $k$ goes large. Meanwhile, some critical points on the left part are focused on. These observations reflect that the module tends to focus on different dimensions of the affinity features at different sampling rates. \wt{We hypothesize that most of the right part is related to motion features, which well explains motion becomes less reliable when the frame rate is lowered. More interestingly, when the frame rate is slightly lower (\eg, $k=8$), the attention map focuses on more different channels, which indicates motion cues are still reliable but it is probably not reliable using the single feature of motion (\eg, IoU). }

\section{Conclusion and Future Work}

In this paper, we introduce Frame Rate Agnostic Multi-Object-Tracking (FraMOT) as an extended challenge of the classical MOT task to seek a wiser solution to the tracking problem. We also present our initial attempt to address the new challenge via the Frame Rate Agnostic MOT Framework with a Periodic training Scheme (FAPS). In the proposed framework, the carefully designed Frame Rate Agnostic Association Module (FAAM) is able to infer frame rate information automatically and utilize it to aid the identity matching of more complicated motion-appearance relations; while the Periodic Training Scheme (PTS) aligns the training and inference environment and help reduce the enlarged gap between training and inference in FraMOT. The two proposed approaches successfully avoid the tracking performance from dropping dramatically when the input frame rate changes, and help build a more robust tracking algorithm. Experimental results and analysis have shown that the proposed methods are effective.

In the future, we may continue to focus on how to further improve the tracking stability against various frame rates. The core problem might be how to develop better frame rate perception methods to fully utilize the video information to aid the task. Besides, FraMOT makes it more difficult to utilize tricky strategies for obtaining improvement.  We may find more robust strategies that work on multiple frame rates rather than work on the normal frame rate only, and thus improve the tracking robustness.

\backmatter

% \bmhead{Supplementary information}

% If your article has accompanying supplementary file/s please state so here. 

% Authors reporting data from electrophoretic gels and blots should supply the full unprocessed scans for key as part of their Supplementary information. This may be requested by the editorial team/s if it is missing.

% Please refer to Journal-level guidance for any specific requirements.

\bmhead{Acknowledgments}
Wanli Ouyang was supported by the Australian Research Council Grant DP200103223, Australian Medical Research Future Fund MRFAI000085, CRC-P Smart Material Recovery Facility (SMRF) – Curby Soft Plastics, and CRC-P ARIA - Bionic Visual-Spatial Prosthesis for the Blind.

\bmhead{Data Availability}
All data used in this paper are publicly available on corresponding websites. 

MOT17/MOT20: \hyperlink{https://motchallenge.net/}{motchallenge.net}; 

CrowdHuman: \hyperlink{https://www.crowdhuman.org/}{www.crowdhuman.org}; 

CityScapes: \hyperlink{https://www.cityscapes-dataset.com/}{www.cityscapes-dataset.com}; 

HIE: \hyperlink{http://humaninevents.org/}{humaninevents.org}.

% \bigskip

% Springer journals and proceedings: \url{https://www.springer.com/gp/editorial-policies}

% Nature Portfolio journals: \url{https://www.nature.com/nature-research/editorial-policies}

% \textit{Scientific Reports}: \url{https://www.nature.com/srep/journal-policies/editorial-policies}

% BMC journals: \url{https://www.biomedcentral.com/getpublished/editorial-policies}
% \end{flushleft}

\newpage
\begin{appendices}

\renewcommand\thefigure{A\arabic{figure}}
\setcounter{figure}{0}
\renewcommand\thetable{A\arabic{table}}
\setcounter{table}{0}

\section{Tacking results Demo}\label{secA1}
\begin{figure}
    \centering
    \includegraphics[width=2.9in]{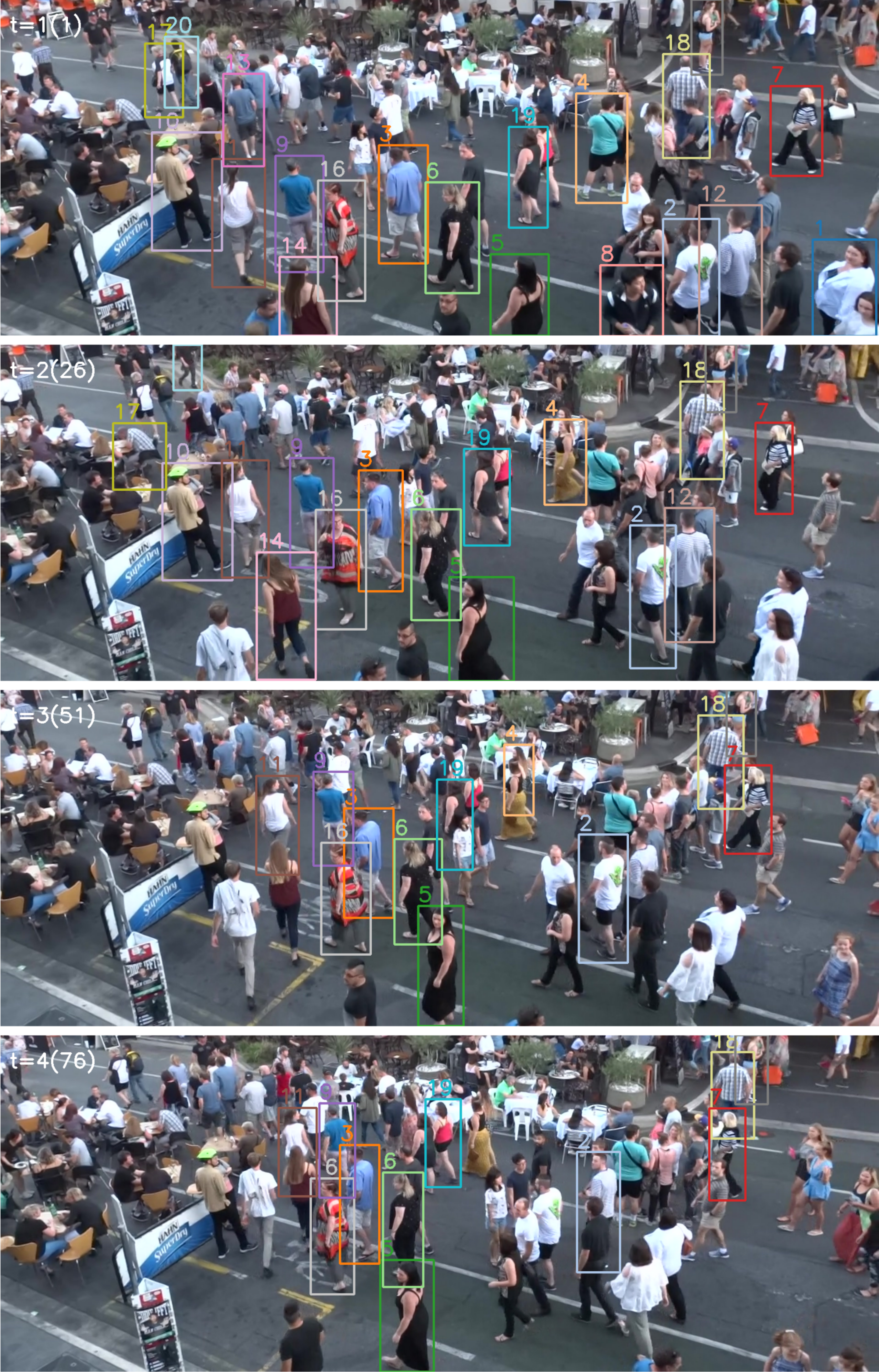}
    \caption{Tracking results on testing sequence MOT20-06, with $k=25$. For better demonstration, only 20 trajectories are shown.}
    \label{fig:trs1}
\end{figure}
\begin{figure}
    \centering
    \includegraphics[width=2.8in]{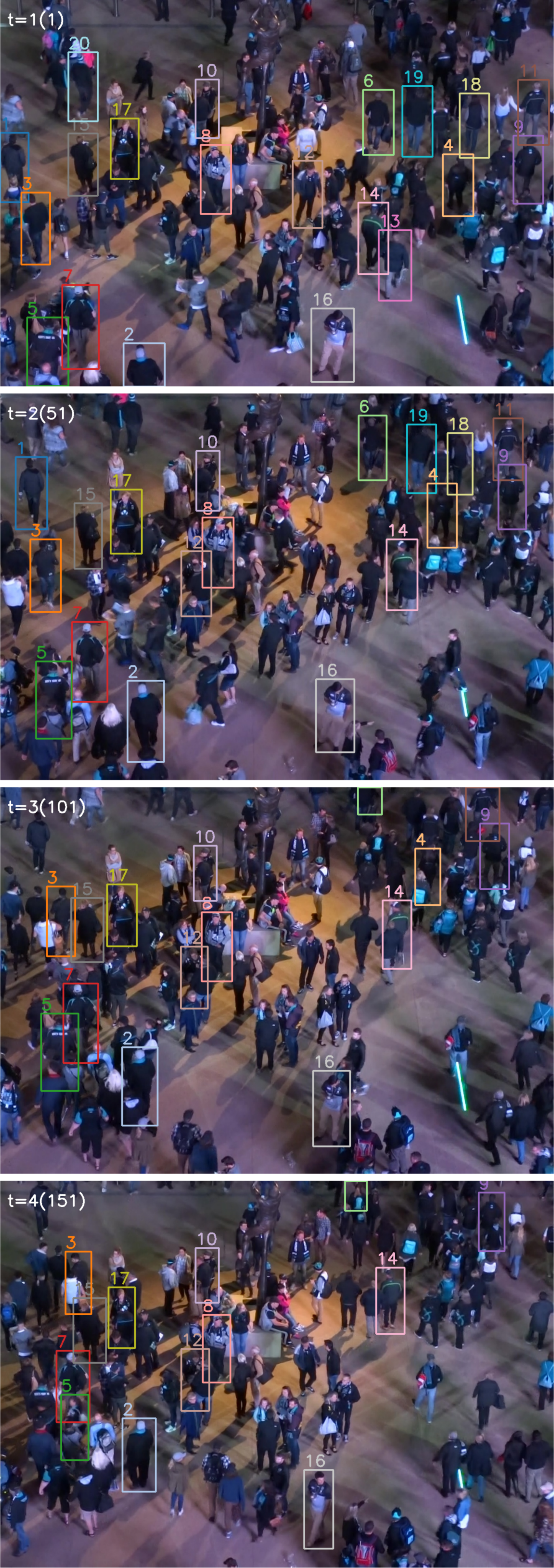}
    \caption{Tracking results on testing sequence MOT20-04, with $k=50$.}
    \label{fig:trs2}
\end{figure}

Fig.~\ref{fig:trs1} and Fig.~\ref{fig:trs2} show some selected tracking results of our approach on the MOT20 testing set. Different colors and different numbers represent different trajectories. We can see that at the lower frame rate scenarios (\ie, sampling factor $k\geq25$), our method keeps a stable performance. More importantly, the results are all from the same model checkpoint, showing that our method can handle various frame rates robustly. Among these scenarios, the movements of the targets between adjacent frames are much larger than those in normal frame rate scenarios. 

 \begin{table*}
 \small
 \centering
 \begin{tabular}{cc@{\ \ \ }c@{\ \ \ }c@{\ \ \ }c@{\ \ \ }c@{\ \ \ }c@{\ \ \ }c@{\ \ \ }c@{\ \ \ }c}
 \cline{1-5}
 \cline{6-10}
 \centering Methods & mHOTA$\uparrow$ & mMOTA$\uparrow$ & mIDF1$\uparrow$ & VR$\downarrow$ \\
 \cline{1-10}
ByteTrack(\cite{zhang2021bytetrack}) & 50.4 & 53.5 & 60.4 & 27.4 \\
FairMOT(\cite{zhang2021fairmot}) & 34.5 & 23.1 & 42.3 & 17.2 \\
GSDT(\cite{wang2021joint}) & 30.4 & 26.2 & 37.7 & 14.7 \\
Ours & \best{54.5} & \best{62.9} & \best{66.8} & \best{10.2} \\ \cline{1-5}
    \end{tabular}
    \caption{Performance on SOMPT22 validation set, unknown frame rate mode. \best{Bold} indicates the best results.}
    \label{tab:sompt22}
\end{table*}

In Fig.~\ref{fig:trs2}, the targets of number 2, number 6 and number 19 do not have a large bounding box overlap between adjacent frames, leading to less reliable motion cues. In normal frame rate scenarios, such a large motion gap in adjacent frames usually indicates a different identity, which is quite different from lower frame rate scenarios. Thanks to the FAAM design and PTS strategy, Our method is able to make a correct prediction in multi-frame-rate settings simultaneously.

% An appendix contains supplementary information that is not an essential part of the text itself but which may be helpful in providing a more comprehensive understanding of the research problem or it is information that is too cumbersome to be included in the body of the paper.





\section{Additional Experiments on SOMPT22}\label{secA2}

Table~\ref{tab:sompt22} shows the results of some recent SOTA methods and our approach on the multiple frame rate version of a new MOT dataset called SOMPT22~\citep{simsek2023sompt22}. SOMPT22 consists of 9 videos in the training set and 5 videos in the testing set. This new dataset has been created to enhance tracking annotations with videos of middle-level densities and static camera views. Compared to MOT20, which also comprises static camera views, SOMPT22 encompasses a wider range of scenarios with various object motions. All methods have used the same joint extractor configuration and checkpoints while testing on MOT20. We have used the same method as MOT17/20 splitting for dividing the SOMPT22 training set and creating the validation set. The reported outcomes have been tested on the validation set.

From Table~\ref{tab:sompt22}, it is evident that our method continues to outperform the recent SOTA methods. Compared to the ByteTrack approach, we have gained improvements of 4.1\%, 9.4\%, and 6.4\% in terms of mHOTA, mMOTA, and mIDF1, respectively. It is noteworthy that both ByteTrack and our method have a deeper backbone. Therefore, when a new dataset with vastly different scenarios is considered, these two methods perform better, while FairMOT and GSDT are less capable of transferring their detection and tracking ability to new environments.

\end{appendices}

\newpage
\bibliography{sn-bibliography}% common bib file

\begin{thebibliography}{37}
\providecommand{\natexlab}[1]{#1}
\providecommand{\url}[1]{{#1}}
\providecommand{\urlprefix}{URL }
\providecommand{\doi}[1]{\url{https://doi.org/#1}}
\providecommand{\eprint}[2][]{\url{#2}}
 \bibcommenthead

\bibitem[{Bernardin and Stiefelhagen(2008)}]{bernardin2008evaluating}
Bernardin K, Stiefelhagen R (2008) Evaluating multiple object tracking
  performance: the clear mot metrics. Journal on Image and Video Processing
  2008:1

\bibitem[{Bras{\'o} and Leal-Taix{\'e}(2020)}]{braso2020learning}
Bras{\'o} G, Leal-Taix{\'e} L (2020) Learning a neural solver for multiple
  object tracking. In: Proceedings of the IEEE/CVF Conference on Computer
  Vision and Pattern Recognition, pp 6247--6257

\bibitem[{Chu and Ling(2019)}]{Chu_2019_ICCV}
Chu P, Ling H (2019) Famnet: Joint learning of feature, affinity and
  multi-dimensional assignment for online multiple object tracking. In: The
  IEEE International Conference on Computer Vision (ICCV)

\bibitem[{Chu et~al(2017)Chu, Ouyang, Li, Wang, Liu, and Yu}]{chu2017online}
Chu Q, Ouyang W, Li H, et~al (2017) Online multi-object tracking using
  cnn-based single object tracker with spatial-temporal attention mechanism.
  In: ICCV

\bibitem[{Cordts et~al(2016)Cordts, Omran, Ramos, Rehfeld, Enzweiler, Benenson,
  Franke, Roth, and Schiele}]{Cordts2016Cityscapes}
Cordts M, Omran M, Ramos S, et~al (2016) The cityscapes dataset for semantic
  urban scene understanding. In: Proc. of the IEEE Conference on Computer
  Vision and Pattern Recognition (CVPR)

\bibitem[{Dendorfer et~al(2020)Dendorfer, Rezatofighi, Milan, Shi, Cremers,
  Reid, Roth, Schindler, and Leal-Taixé}]{dendorfer2020mot20}
Dendorfer P, Rezatofighi H, Milan A, et~al (2020) Mot20: A benchmark for multi
  object tracking in crowded scenes. \eprint{2003.09003}

\bibitem[{Dendorfer et~al(2021)Dendorfer, Osep, Milan, Schindler, Cremers,
  Reid, Roth, and Leal-Taix{\'e}}]{dendorfer2021motchallenge}
Dendorfer P, Osep A, Milan A, et~al (2021) Motchallenge: A benchmark for
  single-camera multiple target tracking. International Journal of Computer
  Vision 129(4):845--881

\bibitem[{Ge et~al(2021)Ge, Liu, Wang, Li, and
  Sun}]{DBLP:journals/corr/abs-2107-08430}
Ge Z, Liu S, Wang F, et~al (2021) {YOLOX:} exceeding {YOLO} series in 2021.
  CoRR abs/2107.08430. \urlprefix\url{https://arxiv.org/abs/2107.08430},
  {\href{https://arxiv.org/abs/2107.08430}{{https://arxiv.org/abs/2107.08430}}}

\bibitem[{Han et~al(2022)Han, Bai, Gao, Wang, and Ouyang}]{han2022dr}
Han T, Bai L, Gao J, et~al (2022) Dr. vic: Decomposition and reasoning for
  video individual counting. In: Proceedings of the IEEE/CVF Conference on
  Computer Vision and Pattern Recognition, pp 3083--3092

\bibitem[{Hornakova et~al(2020)Hornakova, Henschel, Rosenhahn, and
  Swoboda}]{hornakova2020lifted}
Hornakova A, Henschel R, Rosenhahn B, et~al (2020) Lifted disjoint paths with
  application in multiple object tracking. In: International conference on
  machine learning, PMLR, pp 4364--4375

\bibitem[{Hu et~al(2020)Hu, Shi, Zhou, Xing, Ling, and Maybank}]{hu2020dual}
Hu W, Shi X, Zhou Z, et~al (2020) Dual l1-normalized context aware tensor power
  iteration and its applications to multi-object tracking and multi-graph
  matching. International Journal of Computer Vision 128(2):360--392

\bibitem[{Kieritz et~al(2018)Kieritz, Hubner, and Arens}]{kieritz2018joint}
Kieritz H, Hubner W, Arens M (2018) Joint detection and online multi-object
  tracking. In: CVPRW

\bibitem[{Li et~al(2020)Li, Gao, and Jiang}]{li2020graph}
Li J, Gao X, Jiang T (2020) Graph networks for multiple object tracking. In:
  Proceedings of the IEEE/CVF Winter Conference on Applications of Computer
  Vision, pp 719--728

\bibitem[{Lin et~al(2020)Lin, Liu, Liu, Li, Qi, Qian, Wang, Sebe, Xu, Xiong,
  and Shah}]{lin2020human}
Lin W, Liu H, Liu S, et~al (2020) Human in events: A large-scale benchmark for
  human-centric video analysis in complex events. \eprint{2005.04490}

\bibitem[{Luiten et~al(2021)Luiten, Osep, Dendorfer, Torr, Geiger,
  Leal-Taix{\'e}, and Leibe}]{luiten2021hota}
Luiten J, Osep A, Dendorfer P, et~al (2021) Hota: A higher order metric for
  evaluating multi-object tracking. International journal of computer vision
  129(2):548--578

\bibitem[{Ma et~al(2021)Ma, Yang, Li, Jia, Xie, and Gao}]{ma2021deep}
Ma C, Yang F, Li Y, et~al (2021) Deep trajectory post-processing and position
  projection for single \& multiple camera multiple object tracking.
  International Journal of Computer Vision 129(12):3255--3278

\bibitem[{Maksai and Fua(2019)}]{maksai2019eliminating}
Maksai A, Fua P (2019) Eliminating exposure bias and metric mismatch in
  multiple object tracking. In: Proceedings of the IEEE/CVF Conference on
  Computer Vision and Pattern Recognition, pp 4639--4648

\bibitem[{Milan et~al(2016)Milan, Leal-Taix\'{e}, Reid, Roth, and
  Schindler}]{MOT16}
Milan A, Leal-Taix\'{e} L, Reid I, et~al (2016) {MOT}16: {A} benchmark for
  multi-object tracking. arXiv:160300831 [cs]
  \urlprefix\url{http://arxiv.org/abs/1603.00831}, arXiv: 1603.00831

\bibitem[{Milan et~al(2017)Milan, Rezatofighi, Dick, Reid, and
  Schindler}]{milan2017online}
Milan A, Rezatofighi SH, Dick AR, et~al (2017) Online multi-target tracking
  using recurrent neural networks. In: AAAI

\bibitem[{Ristani et~al(2016)Ristani, Solera, Zou, Cucchiara, and
  Tomasi}]{ristani2016performance}
Ristani E, Solera F, Zou R, et~al (2016) Performance measures and a data set
  for multi-target, multi-camera tracking. In: European Conference on Computer
  Vision, Springer, pp 17--35

\bibitem[{Sadeghian et~al(2017)Sadeghian, Alahi, and
  Savarese}]{sadeghian2017tracking}
Sadeghian A, Alahi A, Savarese S (2017) Tracking the untrackable: Learning to
  track multiple cues with long-term dependencies. In: ICCV

\bibitem[{Saleh et~al(2021)Saleh, Aliakbarian, Rezatofighi, Salzmann, and
  Gould}]{saleh2021probabilistic}
Saleh F, Aliakbarian S, Rezatofighi H, et~al (2021) Probabilistic tracklet
  scoring and inpainting for multiple object tracking. In: Proceedings of the
  IEEE/CVF Conference on Computer Vision and Pattern Recognition, pp
  14,329--14,339

\bibitem[{Shao et~al(2018)Shao, Zhao, Li, Xiao, Yu, Zhang, and
  Sun}]{shao2018crowdhuman}
Shao S, Zhao Z, Li B, et~al (2018) Crowdhuman: A benchmark for detecting human
  in a crowd. arXiv preprint arXiv:180500123

\bibitem[{Takala and Pietikainen(2007)}]{takala2007multi}
Takala V, Pietikainen M (2007) Multi-object tracking using color, texture and
  motion. In: 2007 IEEE Conference on Computer Vision and Pattern Recognition,
  IEEE, pp 1--7

\bibitem[{Tang et~al(2016)Tang, Andres, Andriluka, and Schiele}]{tang2016multi}
Tang S, Andres B, Andriluka M, et~al (2016) Multi-person tracking by multicut
  and deep matching. In: ECCV

\bibitem[{Tang et~al(2017)Tang, Andriluka, Andres, and
  Schiele}]{tang2017multiple}
Tang S, Andriluka M, Andres B, et~al (2017) Multiple people tracking by lifted
  multicut and person reidentification. In: CVPR

\bibitem[{Wang et~al(2021)Wang, Kitani, and Weng}]{wang2021joint}
Wang Y, Kitani K, Weng X (2021) Joint object detection and multi-object
  tracking with graph neural networks. In: 2021 IEEE International Conference
  on Robotics and Automation (ICRA), IEEE, pp 13,708--13,715

\bibitem[{Wen et~al(2017)Wen, Lei, Chang, Qi, and Lyu}]{wen2017multi}
Wen L, Lei Z, Chang MC, et~al (2017) Multi-camera multi-target tracking with
  space-time-view hyper-graph. International Journal of Computer Vision
  122(2):313--333

\bibitem[{Wojke et~al(2017)Wojke, Bewley, and Paulus}]{wojke2017simple}
Wojke N, Bewley A, Paulus D (2017) Simple online and realtime tracking with a
  deep association metric. In: ICIP

\bibitem[{Wu et~al(2021)Wu, Cao, Song, Wang, Yang, and Yuan}]{wu2021track}
Wu J, Cao J, Song L, et~al (2021) Track to detect and segment: An online
  multi-object tracker. In: Proceedings of the IEEE/CVF conference on computer
  vision and pattern recognition, pp 12,352--12,361

\bibitem[{Xu et~al(2019)Xu, Cao, Zhang, and Hu}]{xu2019spatial}
Xu J, Cao Y, Zhang Z, et~al (2019) Spatial-temporal relation networks for
  multi-object tracking. In: ICCV

\bibitem[{Yoon et~al(2019)Yoon, Lee, Yang, and Yoon}]{yoon2019structural}
Yoon JH, Lee CR, Yang MH, et~al (2019) Structural constraint data association
  for online multi-object tracking. International Journal of Computer Vision
  127(1):1--21

\bibitem[{Yu et~al(2016)Yu, Li, Li, Liu, Shi, and Yan}]{yu2016poi}
Yu F, Li W, Li Q, et~al (2016) Poi: Multiple object tracking with high
  performance detection and appearance feature. In: ECCV

\bibitem[{Zhang et~al(2008)Zhang, Li, and Nevatia}]{zhang2008global}
Zhang L, Li Y, Nevatia R (2008) Global data association for multi-object
  tracking using network flows. In: CVPr

\bibitem[{Zhang et~al(2021)Zhang, Wang, Wang, Zeng, and Liu}]{zhang2021fairmot}
Zhang Y, Wang C, Wang X, et~al (2021) Fairmot: On the fairness of detection and
  re-identification in multiple object tracking. International Journal of
  Computer Vision 129(11):3069--3087

\bibitem[{Zhang et~al(2022)Zhang, Sun, Jiang, Yu, Weng, Yuan, Luo, Liu, and
  Wang}]{zhang2021bytetrack}
Zhang Y, Sun P, Jiang Y, et~al (2022) Bytetrack: Multi-object tracking by
  associating every detection box. In: Proceedings of the European Conference
  on Computer Vision (ECCV)

\bibitem[{Zhou et~al(2020)Zhou, Koltun, and
  Kr{\"a}henb{\"u}hl}]{zhou2020tracking}
Zhou X, Koltun V, Kr{\"a}henb{\"u}hl P (2020) Tracking objects as points. ECCV

\end{thebibliography}


\end{document}